\documentclass[journal]{IEEEtran}

\makeatletter
\def\BState{\State\hskip-\ALG@thistlm}
\makeatother
\usepackage{color}
\usepackage{epsf}
\usepackage{times}
\usepackage{epsfig}
\usepackage{epstopdf}
\usepackage{graphicx}
\usepackage{algorithm}
\usepackage{amssymb}
\usepackage{amsxtra}
\usepackage{multirow}
\usepackage{mathtools}
\usepackage{multirow}
\usepackage{comment}
\usepackage[normalem]{ulem}
\usepackage{tabu}
\usepackage[dvipsnames]{xcolor}
\usepackage{multicol}
\usepackage{cite}
\usepackage{caption}
\usepackage{subfigure}
\DeclareMathOperator*{\argmax}{arg\,max}
\begin{document}

\vspace{-0.0cm}
\title{Low Complexity Recruitment for Collaborative Mobile Crowdsourcing Using Graph Neural Networks}
\author{
\IEEEauthorblockN{Aymen Hamrouni,  \textit{Student Member, IEEE},  Hakim Ghazzai, \textit{Senior Member, IEEE}, Turki Alelyani, \textit{Member, IEEE},  and Yehia Massoud, \textit{Fellow, IEEE}\vspace{-0.0cm}}\\
{\thanks {\hrule
\vspace{0.1cm} 
\indent A part of this work has been accepted for publication in the proceedings of IEEE Technology \& Engineering Management Conference, Novi, MI, USA, June. 2020~\cite{TEMSCON2020}.

Aymen Hamrouni, Hakim Ghazzai, and Yehia Massoud are with the School of Systems and Enterprises, Stevens Institute of Technology, Hoboken, NJ, USA. (E\textendash mails: \{ahamroun, hghazzai, ymassoud\}@stevens.edu).\newline
Turki Alelyani is with the College of Computer Science and Information Systems, Najran University, Najran, Saudi Arabia, (E\textendash mail: tnalelyani@nu.edu.sa).\newline
}}\vspace{-0.0cm}}

\maketitle
\thispagestyle{empty}

\begin{abstract}
\boldmath{
Collaborative Mobile crowdsourcing (CMCS) allows entities, e.g., local authorities or individuals, to hire a team of workers from the crowd of connected people, to execute complex tasks. In this paper, we investigate two different CMCS recruitment strategies allowing task requesters to form teams of socially connected and skilled workers: i) a platform-based strategy where the platform exploits its own knowledge about the workers to form a team and ii) a leader-based strategy where the platform designates a group leader that recruits its own suitable team given its own knowledge about its Social Network (SN) neighbors.  We first formulate the recruitment as an Integer Linear Program (ILP) that optimally forms teams according to four fuzzy-logic-based criteria: level of expertise, social relationship strength, recruitment cost, and recruiter's confidence level. To cope with NP-hardness, we design a novel low-complexity CMCS recruitment approach relying on Graph Neural Networks (GNNs), specifically graph embedding and clustering techniques, to shrink the workers’ search space and afterwards, exploiting a meta-heuristic genetic algorithm to select appropriate workers. Simulation results applied on a real-world dataset illustrate the performance of both proposed CMCS recruitment approaches. It is shown that our proposed low-complexity GNN-based recruitment algorithm achieves close performances to those of the baseline ILP with significant computational time saving and ability to operate on large-scale mobile crowdsourcing platforms. It is also shown that compared to the leader-based strategy, the platform-based strategy recruits a more skilled team but with lower SN relationships and higher cost.}
\end{abstract}

\vspace{-0.0cm}
\section{Introduction}
\label{Sec1a}

Mobile Crowdsourcing (MCS) is the act of outsourcing sensing tasks traditionally performed by employees or contractors to an undefined large group of dynamic Internet population or cyber community through an open or targeted call. It harnesses the power of built-in sensors in mobile devices (e.g., smartphones, tablets, and smart devices) and allows Internet-of-things (IoT) devices to establish relationships, communicate, and cooperate together to complete specific sensing and data collection tasks without requiring pre-deployed dedicated infrastructure~\cite{9119125,9102392}.

In the traditional MCS architecture~\cite{8995775}, three main components are identified: task requesters, task workers, and the Platform as a Service (PaaS) hosting the main framework. The task requester, acting as a human possessing a smartphone or an IoT device (e.g., ground or aerial autonomous vehicle), provides its task description and criteria to the PaaS server requesting a certain service, e.g., covering an ongoing event by taking pictures~\cite{8982179}. The cloud platform then uses these requirements to match suitable workers and provides them with the task information. 
For this paradigm, the selected workers are asked to achieve what is necessary independently of each other and the final result is combined from their partial results to produce the overall outcome. The quality of the returned result, which is deduced from each worker's partial result, depends heavily on the characteristics of the recruited workers. To this end, most of the MCS research focused on hiring professional workers for each task such that they can complete the tasks' successfully and provide desired results~\cite{8884949,8672075}.

Nevertheless, in many other MCS applications, the set of tasks, also called projects, are very complex and the success of their completion depends not only on the expertise of their selected workers but also on how efficiently these workers can work together as a team. This could be, for example, the case of a MCS framework that helps build a virtual search party of smartphone users to find lost items, pets, or persons, as well as returning them. This team-based MCS paradigm is referred to as Collaborative MCS (CMCS)~\cite{8633829}. The enlisted workers are divided into convenient groups, using specific criteria required by the requester, and are asked to collaborate for the search by providing up-to-the minute reports about any updates. If the collaboration somehow fails for one reason or another, the job cannot be achieved successfully. Therefore, besides providing the required skills, the successful completion of the project is very sensitive to the way team members collaborate and communicate.

In this paper, we develop a team recruitment framework for large-scale CMCS systems. The objective is to recruit the most suitable workers that will collaborate and complete a CMCS project while considering various workers’ and tasks’ constraints and attributes. The proposed CMCS recruitment framework uses fuzzy logic for both the workers’ skills and Social Network (SN) relations connecting the different workers registered in the CMCS platform. For an appropriate matching, it aims to maximize a multi-objective function enclosing four key recruitment metrics: select skilled workers that meet the project requirements, select socially connected team, reduce the total recruitment cost, and ensure a high confidence level of the team selection process.  

Two recruitment strategies are investigated: The first one is a platform-based strategy in which the CMCS platform itself is responsible for forming the entire team based on its knowledge about the workers' SN and their attributes (e.g., profile, history, experience, previous performance, reliability). The second one is a leader-based strategy in which the cloud platform selects a worker as a leader to which it delegates the team formation procedure. The chosen leader recruits team members based on its knowledge about other workers in its SN vicinity (e.g., social incentive mechanism, man-to-man, friendship). For both strategies, we model the CMCS recruitment problem as an Integer Linear Program (ILP) with the aim to recruit optimal teams. However, this comes at the expense of a very high computational complexity. Therefore, and in order to promote instantaneous real-time recruitment in large-scale CMCS platforms, we propose a low-complexity recruitment approach that relies on Graph Neural Networks (GNNs)~\cite{8294302,4703190} to learn unsupervised representations of the workers' relational structure and attributes by using graph embedding techniques as a way to reduce the complexity due to the problem's dimensionality~\cite{9184643}. Afterwards, for the resulting reduced search space, a modified meta-heuristic algorithm, namely the Genetic Algorithm (GA), is employed to form teams that match the CMCS project requirements.

In layman’s terms, the main contributions of this paper are summarized as follows:\\
$\bullet$ Formulate an ILP optimization problem that optimally solves the CMCS recruitment problem using two recruitment strategies: i) platform-based recruitment strategy that exploits the knowledge of the platform towards workers and recruits suitable team members and ii) a leader-based recruitment strategy that uses the knowledge of an appointed leader by the platform to recruit the rest of the team. \\
$\bullet$ Design a low-complexity GNN-based algorithm that maps workers' attributes and representations from their SN graph into a 2-D format then clusters workers based their attributes and computes locally an approximation GA to form teams.

Simulation results on the Facebook SN dataset illustrate the performance of the CMCS framework for selected scenarios and show that our proposed CMCS recruitment algorithm for specific embedding parameters achieves close results to the baseline ILP with significant computational time saving. It is also shown that the proposed approach outperforms existing algorithms in terms of team recruitment and computational speed. Further experimental results show a performance trade-off between the two virtual team grouping strategies when varying the members SN edge degrees. Compared to the leader-based strategy, the platform-based strategy recruits a more skilled team but with lower SN relationships and higher cost.

The remainder of this paper is organized as follows. Related work is reviewed in Section~\ref{RelatedWork}. The system model is presented in Section~\ref{Sec2}. The ILP formulation for solving the CMCS recruitment problem is introduced in Section~\ref{sec3}. Section~\ref{sec4} presents the proposed low-complexity CMCS recruitment algorithm based on Graph Embedding techniques. Experiments and simulation results are discussed in Section~\ref{sec5}. Finally, the paper is concluded in Section~\ref{sec6}.
\section{Related Work}
\label{RelatedWork}

\textcolor{black}{Over the last decade, many MCS-based applications have been emerging. Zheng et al.~\cite{Zheng:2016:DDC:3025111.3025118} examined user skills using Knowledge Base (KB), e.g., Wikipedia and Freebase, to detect the domain of tasks and workers. Gong et al.~\cite{7365407} discussed privacy issues when recommending tasks to workers. \textcolor{black}{In this context, the authors of~\cite{DBLP:journals/corr/abs-2011-04345} have proposed a joint Bayesian learning and graph optimization framework achieving privacy-preserving and fast convergence over a heterogeneous and
sparse IoT graph without assumptions on the prior knowledge of the data distribution across the nodes.} Yan et al.~\cite{8835905} proposed an algorithm to enhance High Definition (HD) and Ultra High Definition (UHD) video services for mobile users through MCS participants that download and transmit video segments. Hamrouni et al.~\cite{8884949} proposed a photo-based MCS architecture that monitors the quality of the photos submitted by reporters to cover ongoing events. Alabduljabbar et al.~ \cite{8672075} studied a dynamic approach that exploits task-quality ontology to select the most suitable quality control mechanism for a given task based on its type. Ben Said et al.~\cite{8744231} proposed an algorithm to determine the best public transport journey plan offering based on the quality of service of available WiFi mobile crowdsourced WiFi coverage along the journey.}

\textcolor{black}{In the MCS applications, the requester posts tasks and recruits workers to complete them independently of each other~\cite{9138420}. Due to the temporal and spatial constraints of the tasks and background knowledge of different workers, not everyone is qualified to execute a specific task. Therefore, selecting the appropriate worker from a pool of candidates is important as it can directly affect the quality of the final result of the MCS task~\cite{8338433}. Estrada et al. \cite{7974784} proposed a MCS recruitment algorithm that implements a multi-objective task allocation algorithm based on Particle Swarm Optimization (PSO) while handling incoming sensing tasks in the server side and at the end-user side and avoiding delaying or declining the sensing requests due to unforeseen user context. Guo et al.~\cite{8316812} reviewed the state of research in many types of task allocation such as multiple task allocation, low-cost task allocation, etc. They discussed the recruitment issues in real-world deployment but proposed solutions in a small-scale problem. Gao et al.~\cite{9016107} proposed a Learning-based Credible Participant Recruitment Strategy (LC-PRS) to maximize the platform and participants' profits at the same time via MCS participation. Yu et al.~\cite{8666714} proposed a dynamic utility task allocation algorithm that estimates the ability of workers to execute tasks and then maximizes the overall utility. Xing et al.~\cite{8834775} proposed an intelligent multi-attribute crowdsourcing task assignment approach. This approach solves the stability of the crowdsourcing transaction and adopts the game theory to maximize the satisfaction of both sides during the task assignment. Yu et al.~\cite{8667429} developed an approach to recruit suitable workers using three metrics: worker's ability, task module complexity, and worker's active time. Wang et al. \cite{8434340} proposed to hire MCS workers by leveraging the influence of information propagation on the social networks.
Other studies have focused on the optimization of spatial task assignment. For instance, Zhou et al.~\cite{9177035} proposed a secure and efficient spatial task matching framework that utilizes multi-user searchable encryption and secure index technique. Abdullah et al.~\cite{9127467} introduced the use of Bayesian Network in modelling and selecting optimal workers and used $k$-medoids partitioning technique for tasks clustering and scheduling. Wang et al.~\cite{9103617} put the SMCS into another prospective by formulating it as the Walrasian equilibrium where the optimum solution is researched to maximize the social welfare of mobile crowdsourcing systems. Hamrouni et al.~\cite{9028164} proposed a heuristic SMCS recruitment approach allowing the achievement of sub-optimal matching and recruitment solution by iteratively solving a weighted bipartite graph problem.}
\begin{figure}[t]
%\begin{minipage}[h]{1\linewidth}
    \centering
    \vspace{0.1cm}
    \frame{\includegraphics[width=8.5cm]{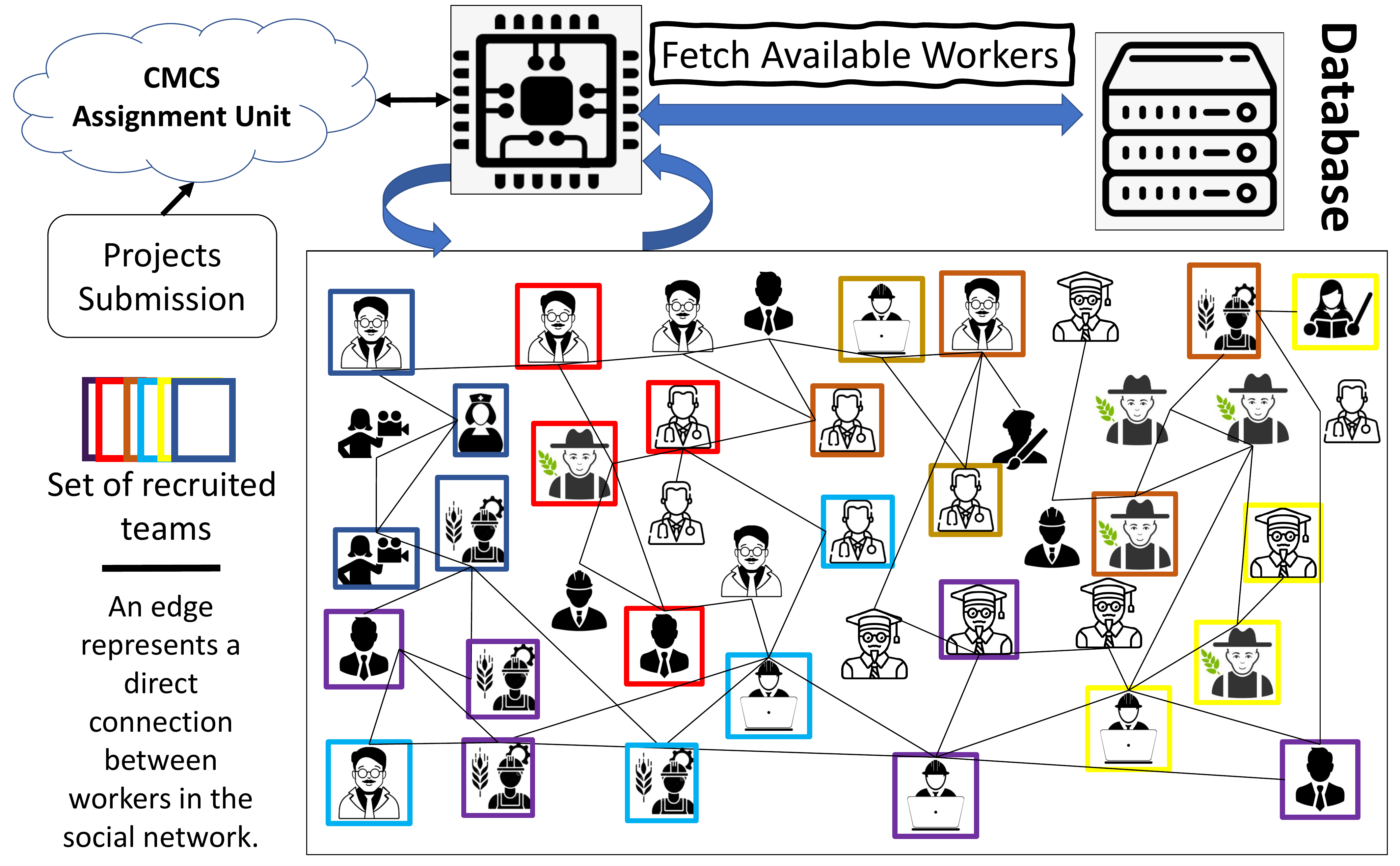}}
    \caption{Recruitment system model in the CMCS Framework displaying samples of team workers and their respective teams.}
    \label{figure1}\vspace{-0.5cm}
%\end{minipage}
\end{figure}

On the other hand, as a teamwork-based paradigm, CMCS aims to form groups of workers, often with diverse and complementary skills, that will work together to complete complex tasks~\cite{2222,7806269}. Fewer studies have begun to address the need to consider recruiting a team of workers in MCS~\cite{8423113,7248382,7070400}. Some approaches, such as~\cite{116}, focused on dividing complex tasks into flows of simple sub-tasks and allocating these sub-tasks to a team of workers. These approaches have focused only on the expertise of recruited team members and did not consider the interaction within members. Other approaches limited their focus on team formation in SNs and proposed a solution to hire teams with good social relationships indifferent of the members' levels of expertise~\cite{8488386,article7,article8}. \textcolor{black}{The authors in~\cite{hamrouni2021collaborative} have shed the light on the CMCS paradigm by providing a detailed CMCS taxonomy and investigating the challenges in designing CMC tasks. }

Most of the existing studies have investigated traditional MCS platforms and offered solutions designed for simple MCS tasks. The ones that are designed for CMCS platforms are focused on either the workers' social network or their attributes and skills. To the best of our knowledge, this work is the first one that tackles the recruitment process for complex projects within CMCS systems while relying on both the workers' attributes and social network. It is also the first that proposes a machine learning approach, specifically a GNN-based algorithm, for recruiting skilled and socially connected teams while taking into account the team compensation cost and the confidence level of the recruitment decision.

\begin{table}[t]
\vspace{-0.3cm}
\begin{center}
\caption{\label{table1} Notations and Descriptions}
\vspace{-0.2cm}
\addtolength{\tabcolsep}{-0pt} \scalebox{1}{ \begin{tabular}{|c || c|}%
  \hline
 \textbf{Notations} & \textbf{Descriptions}\\
  \hline
  \hline
					\multirow{1}{*}{$\mathcal P$} &  Set of projects in the framework, $\mathcal P=\{1, \dots, |\mathcal P|\}$  
					\\
		\hline
		 $\mathbf{S_p}$ & A set representing the skills that project $p$ requires\\
		\hline
		\multirow{2}{*}{$\mathcal W$} &  Set of all workers in the framework,\\
		& $\mathcal W=\{1, \dots, |\mathcal W|\}$ \\	\hline
		\multirow{2}{*}{$C^i_{w,k}$} & Cost, seen by recruiter $i$, demanded by worker $w$\\
		&when providing skill $k$
		\\
		\hline
	\multirow{2}{*}{$\mathbf{C_w}$} & A set representing the cost \\
	& for each skill provided by worker $w$
		\\
						\hline
		\multirow{2}{*}{$S^i_{w,k}$} & Skill level, seen by recruiter $i$, provided by\\ & worker $w$ contributing with skill $k$
		\\
							\hline
		\multirow{2}{*}{$S_{w,k}$} & Real skill set value provided by worker $w$\\
		&when contributing with skill $k$
		\\
				\hline
		\multirow{2}{*}{$\mathbf{\hat{S}_w}$} & A set representing the noised skills\\
		&values for worker $w$  \\
		\hline
		\multirow{2}{*}{$E^i_{w,k}$} & Efficiency of worker $w$ \\
		& recruited by $i$ contributing with skill $k$ \\
			\hline
			$U^i_{w}$ & Confidence level of recruiter $i$ towards worker $w$ \\
		\hline
\end{tabular} }
\end{center}\vspace{-0.5cm}
\end{table}
\section{CMCS System Model}\label{Sec2}
In this section, we start by presenting the different components of the CMCS system. Afterwards, we introduce the SN model and the recruitment utility that the task manager aims to optimize during the CMCS assignment process.

A typical CMCS framework interacts with two main actors: project/task requesters and task workers. The entire platform workflow is divided into a four-stage process:
1) a task initiation stage in which requesters define and submit their projects' requirements to the platform, 
2) a task announcement phase during which the platform uses these requirements in order to select suitable teams and notify them about the project,
3) a task execution stage during which the selected team members collaborate and communicate together to collect and submit requested data, and
4) a response handover stage where the results of the team members are combined and delivered to the project requesters.

In this paper, we focus on the second stage where the CMCS platform target is to recruit suitable and convenient team for the assigned project. In Fig.~\ref{figure1}, we illustrate a high-level workflow of the recruitment process in CMCS platform. The CMCS assignment unit server uses both the project's (e.g., required skills) and worker's attributes (e.g., level of expertise, availability), to determine a suitable matching. As we can notice, the server consults its database where information about the registered workers and their related SN graph exists. The objective of the server is to form teams that match the project requirement accordingly. 
Note that the proposed platform is recruiting a team of workers where each of its member contributes with a different skill. However, we may notice that workers with the same profession are in the same team. This is because we assume that each worker is an expert in his/her primary profession but also have some levels of expertise in other ones.  Next, we present the characteristics and attributes of the different CMCS actors. For clarity, we provide the main notations used throughout the paper in Table~\ref{table1}.

\subsection{Project Model}
The overall set of projects published in the platform is denoted by $\mathcal P$ where $
\mathcal P=\{1, \dots, \dots, |\mathcal P|\}.$
Each task $p \in \mathcal P$ is denoted with a list of attributes. For simplification purposes, let $\{Id_{p}, Tit_{p}, Des_{p}, \mathbf{S_{p}} \}$ be the list of attributes of task $p$ that are defined as follows:\\
$\bullet$ $Id_p$: represents the project's unique identifier.\\
$\bullet$ $Tit_p$ and $Des_p$: represent the task title and the text description of the task (e.g., a task can be described as ``Please contact your teammates in order to help in the ongoing fire in the nearby mall'').\\
$\bullet$ $\mathbf{S_p}$:  This attribute represents the skills or the features that a task needs. Given a set $\mathcal S=\{1,\dots, |S|\}$ of all $S$ possible skills in the system, we define the logical skill quantity for a project $p$ by $Q_p(k),\,k\ \in \mathcal S$ where $Q_p(k)=1$ if the skill $k$ is required by project $p$ and $Q_p(k)=0$, otherwise. Hence, the skills set required by the project $\mathbf{S_p} =\{k \in \mathcal S/Q_p(k)=1\}$. In the case of the previous mentioned project example, $\mathbf{S_p}$ can be a binary vector with all zeros except the skills referring to a doctor, a nurse, a fireman, an IT engineer, a mechanical engineer, and photographer.

\subsection{Worker Model}
\label{workermodel}
A worker is a platform user which offers its services to the system to complete CMCS projects. We denote the set of workers in the platform by $\mathcal W=\{1, \dots, |\mathcal W|\}$. Arriving workers express their readiness to collaborate in projects and provide their restraints to the CMCS platform. Each worker $w \in \mathcal W$ is characterized by a list of attributes. For simplification purposes, let $\{id_w, ent_{w}, lev_w, \mathbf{C_w}, \mathbf{\hat{S}_w}\}$ be that list where its elements are explained as follows:\\
$\bullet$ $id_w$: this term represents the worker's unique identifier. \\
$\bullet$ $ent_w$ and $lev_w$: these attributes are the entering and leaving time of the worker to the system, respectively.\\
$\bullet$ $C_w$: this term refers to the reward that the worker is seeking. To execute a project with skill $k$, a worker $w$ may request a certain cost denoted by~$C_{wk}$. Therefore, $\mathbf{C_w}=\{C_{w1},\dots,C_{w|S|}\}$ \\
$\bullet$ $\mathbf{\hat{S}_w}$: Each worker $w \in \mathcal W$ has a degree of expertise in skill $k \in \mathcal S$ denoted by $S_{wk}$ where $0\leq S_{wk} \leq 1$. The term $S_{wk}$ represents the actual expertise value of skill $k$ that worker $w$ has and it is interpreted as follows: $S_{wk} \leftarrow 1$ means that the worker $w$ is an expert in skill $k$. Otherwise, $S_{wk}\rightarrow 0$.  We assume that a recruiter $i$, which can be a worker as well as the platform itself, does not perfectly know the degree of the skill $k$ of each worker. Instead, it knows an estimated value, referred to as a fuzzy skill set, expressed as follows: $\hat{S}^i_{wk}=S_{wk}+\tilde{S}^i_{w}$, where $\tilde{S}^i_{w}$ is a skill error made by the recruiter $i$ given its knowledge about worker $w$. The parameter $\tilde{S}_{w}$ is a noise modeled as a probability distribution with a variance $U_w$ that reflects the confidence level of the recruiter $i$ towards worker $w$.  Let $\mathbf{\hat{S}_w}=\{\hat{S}_{w1}, \dots, \hat{S}_{w|S|}\}$ be the set of skills provided by worker $w$. We suppose that each recruited worker can only contribute with one required skill. Consequently, for a project $p$ having as a skill set $\mathcal S_p$, the number of team members must be $|\mathcal S_p|$. This approach uses the fuzzy logic in which the truth values of variables may be any real number between 0 and 1 both inclusive. It is employed to handle the concept of partial truth, where the truth value may range between completely true and completely false.

 \subsection{Social Network Model}
The subscribed workers in the platform form a SN modeled as an undirected weighted graph $\mathcal G(\mathcal W,\mathcal E)$. Every vertex of $\mathcal G$ corresponds to a worker $w \in \mathcal W$ while the set of edges $ \mathcal E$ represents the SN relationships between the workers. Initially, we only consider the edges connecting a pairwise of workers that can directly communicate and collaborate and we associate the value $1$ to their weights. Then, the edges between the remaining pairwise of vertices, e.g., $(w,w')$, which are not directly connected are given a weight computed using the shortest number of hops, denoted by $n^{hops}_{ww'}$, needed for one of the pairwise vertices to reach the other. Hence, the graph $\mathcal G$ is converted into a complete graph where all vertices are directly connected and the values of the edges' weights indicate the social relationships levels between each pair of workers. The real values on each edge between two workers $w$ and $w'$ are given as: $R_{ww'}=\frac{1}{1+n^{hops}_{ww'}}$.
\textcolor{black}{In a way similar to $\tilde{S}^{i}_{w}$ and due to the fact that social networks do not capture nor reflect complete information about relationships between the workers, we introduce uncertainty in the social network relations. Furthermore, if a worker "A" is directly connected to two other workers, "B" and "C". The social network will naturally attribute the same value to the links between workers "A" and "B" and workers "A" and "C". However, in practice, this could not be the case since worker "A" can have a stronger relationship with worker "B" than worker "C". Other factors that introduce uncertainty towards the connections in the social network can also include, and not limited to, fake profiles or false links. Therefore,  we assume that a recruiter $i$, which can be a worker as well as the platform itself, does not perfectly know the real relationship between workers. Instead, it knows an estimated value, referred to as: $\hat{R}_{ww'}=R_{ww'}+\frac{(\tilde{R}^i_{w}+\tilde{R}^{i}_{w'})}{2}$, where $\tilde{R}^i_{w}$ and $\tilde{R}^i_{w'}$ represent relationships error made by the recruiter $i$ given its knowledge about workers $w$ and $w'$, respectively.}

The noise $\tilde{R}^{i}_{w}$ is modeled as a probability distribution with a variance $U^{i}_w$ that reflects the confidence level of recruiter $i$ towards worker $w$. If an isolated sub-graph exists, then the weights connecting a node of this sub-graph to other external vertices is a pure noise (i.e., $n_{hops}\rightarrow \infty$).

\vspace{-0.2cm}
 \subsection{Recruitment Utility Model}
In the CMCS system, it is important to establish an effective criterion to guarantee an appropriate recruitment decision. The latter should guarantee a certain correlation between the team members and the projects as well as between the team members themselves such that the formed team matches the project requirements. We define this correlation as the efficiency of workers to perform tasks. The efficiency of worker $w$ to contribute with skill $k$ and chosen by a recruiter $i$ is modeled as a Pareto multi-objective function and is expressed as follows: 
\begin{align}
E^i_{w,k}= \eta_1\frac{ \hat{S}^i_{w,k}}{\bar{S}}  - \eta_2 \frac{ U^{i}_w}{\bar{U}}  
-   \eta_3 \frac{C_{w,k}}{\bar{ C }}  + \frac{\eta_4}{|\mathcal T|-1} \hspace{-0.2cm}\sum_{w'\in \mathcal T \backslash\ \{w\}} \hspace{-0.2cm} \frac{\hat{R}_{ww'}}{\bar{R}}.
\label{1}
\end{align}
The efficiency of the worker is measured by its skill level as seen by the recruiter, the confidence level of the recruiter, the requester cost to execute the task, and its social relation level with other team members. In~\eqref{1}, $\mathcal T$ denotes the set of hired workers in the formed team. The quantities $\bar{X}$ are introduced for normalization purposes so that the four key metrics have the same order of magnitude in the efficiency expression. The weights $\eta_t$, with $t \in \{1,2,3,4\}$ and $\sum_{t=1}^{4} \eta_t=1$, indicate the recruiter's recruitment strategy. For example, situations where the project requester only cares about its task being completed by the workers having the highest skills (i.e., $\eta_1=1$, and $\eta_2, \eta_3, \eta_4$ are set to 0). If the recruiter is looking for a reduced cost-effective team, a higher value of $\eta_3$ is recommended.

\section{Optimal Recruitment Problem Formulation}
\label{sec3}

In this section, we propose a formulation to the CMCS recruitment problem using two possible recruitment strategies depending on the nature of the recruiter: the platform itself or a leader delegated by the platform. Let $\mathcal I$ be the set of possible team recruiters, which can be defined as follows:
\begin{align} \label{gamma}
\mathcal I= 
     \begin{cases}
       \text{\{0\},} &\,\text{if the recruiter is the platform,}\\
      \mathcal W, &\,\text{if the recruiter is a worker.}
     \end{cases} \notag
     \end{align}
We propose to model the recruitment problem as an Integer Linear Program. In the following, we define the different decision variables and the necessary constraints that ensure suitable matching of the defined tasks to the appropriate set of workers. Afterwards, we present the objective function that we seek to maximize, which is based on the correlation function defined in ~\eqref{1}.

\subsection{Decision Variables}

In order to assign to the recruiter the chosen workers for a project $p$ and decide the skill $s_k$ that he/she will contribute with, we introduce a binary decision variable $x^i_{wk}$ defined as follows:
\begin{align} 
x^i_{wk}= &
     \begin{cases}
       \text{1,} &\,\text{if recruiter $i$ selects worker $w$ to contribute}\\
       &\text{in project $p$ with skill $k$,}\\
       \text{0,} &\,\text{otherwise,} \\ 
     \end{cases}\notag\\
 &\hspace{2cm}\forall \, (i,w) \in \mathcal I \times \mathcal W, \forall \, k \in \mathcal S_p. \notag
\end{align}
Depending on the recruitment strategy, the index $i$ can take either the value of $\{0\}$ for the platform-based approach or any value in $\mathcal W$ for the leader-based approach. For the latter approach, the optimizer decides which worker in $\mathcal W$ will be assigned as a leader. Hence, it will determine the leader and the recruited team based on the knowledge and SN of the leader.

Another binary decision variable $v_{ww'}$ is introduced to consider the social relationships between the project teammates. This variable is presented as follows:
\begin{align} 
& v_{ww'}= 
     \begin{cases}
       \text{1,} &\,\text{if workers $w$ and $w'$ are hired within the project,}\\
       \text{0,} &\,\text{otherwise,} \\ 
     \end{cases}\notag\\
 &\hspace{2cm}\forall \, (w,w') \in \mathcal W \times \mathcal W. \notag
     \end{align}
The variable $v_{ww'}$ indicates that all the positive 2-tuple combinations of the chosen team members. Its value can be computed using the following expression:
\begin{align}
&v_{ww'}= \sum_{i \in \mathcal I }\sum_{k \in \mathcal S_p}^{} x^i_{wk} \land \sum_{i \in \mathcal I } \sum_{k \in \mathcal S_p}^{} x^i_{w'k}, \forall \, (w,w') \in \mathcal W \times \mathcal W, 
\end{align}
where the symbol ($\land$) represents the logical operator \textit{AND}.

For the leader-based approach, we introduce an endogenous binary variable presented as follows:
\begin{align} \label{v}
& y^{i}= 
     \begin{cases}
       \text{1,} &\,\text{if worker $i$ is selected as a team leader,}\\
       \text{0,} &\,\text{otherwise,} \\ 
     \end{cases}\notag\\
 &\hspace{2cm}\forall \, i \in \mathcal W. \notag
     \end{align}

\subsection{Recruitment Problem Constraints}
\subsubsection{Common Constraints}
\begin{figure*}[t]
\begin{minipage}[h]{1\linewidth}
    \centering
  \frame{  \includegraphics[width=18cm]{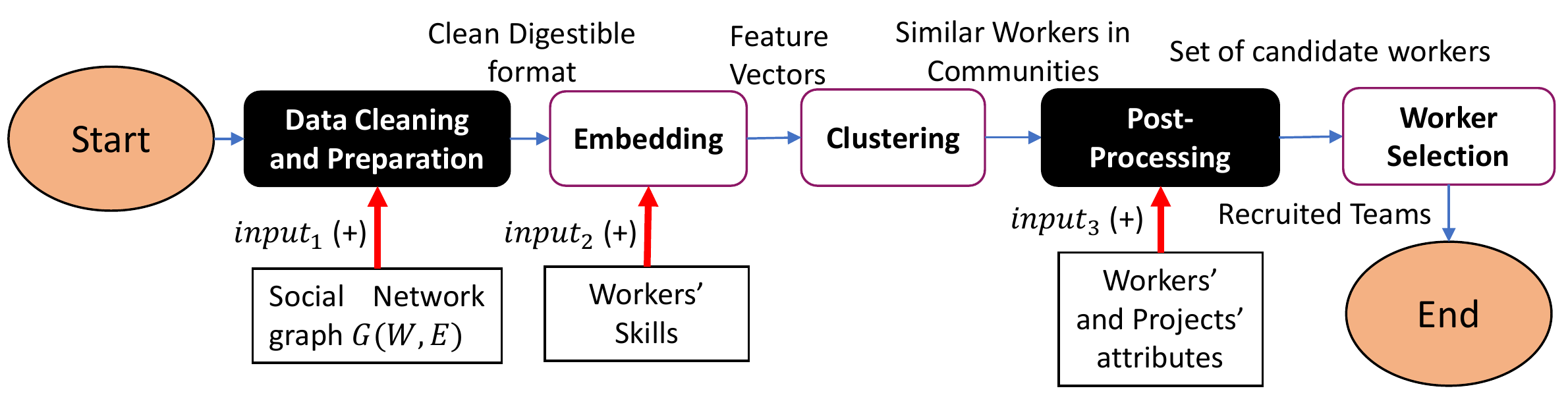}}
    \caption{High-level flowchart illustrating the inputs/outputs of each phase of the proposed CMCS recruitment algorithm.}
    \label{highlevelchart}
\end{minipage}
\end{figure*}

In the following, we present all the common constraints for both recruitment approaches.\\
$\bullet$ \textbf{Skill Participation Constraints:} The following constraint forces each worker, if selected, to provide at most one skill for the project $p$:
%~\footnoteautorefname[5]{
%For readability, we use the following notations \,&\forall \, t, \,&\forall \, k, and \,&\forall \, w to denote \,&\forall \, t,\dots,T \,&\forall \, k,\dots,T and \,&\forall \, w,\dots,W respectively. Otherwise, the range of each index will be specified.
%}
\begin{align}
\sum_{i \in \mathcal I } \sum_{k \in \mathcal S_p}^{}x^i_{wk}\leq 1, \, \forall \, w \in \mathcal W.
\end{align}
$\bullet$ \textbf{Social Network Constraints:} These constraints address the SN of the workers within the team. The following one eliminates the case of counting a worker $w$ to be its own co-worker if hired within the team:
\begin{align}
v_{ww} = 0, \forall \, w \in \mathcal W,  \label{4} 
\end{align}
while the following one forces the symmetric relations between vertices in the undirected graph $\mathcal G(\mathcal W, \mathcal E)$ of workers' SNs:
\begin{align}
v_{w'w} = v_{ww'}, \,   \forall \, (w,w') \in \mathcal W \times \mathcal W.  \label{5} 
\end{align}

\subsubsection{Platform-based Approach Constraints}
This constraint targets the platform-based approach:\\
$\bullet$ \textbf{Capacity Constraints:}
The following constraint ensures that each of the workers within the hired team contributes with a required skill defined by the project.
\begin{equation}
\hspace{-0.4cm}\sum_{i \in \mathcal I } \sum_{w  \in \mathcal W}^{}x^i_{wk} = Q(s_k), \, \forall \, k \in \mathcal S_p,\label{6}
\end{equation}

\subsubsection{Leader-based Approach Constraints}
These constraints target the leader-based approach. \\
$\bullet$ \textbf{Capacity Constraints:}
The following constraint ensures that each of the workers within the hired team contributes with a required skill defined by the project.
\begin{subequations}
\begin{align}
&\sum_{k \in \mathcal S_p}^{} \left( \sum_{w  \in \mathcal W}^{} x^i_{wk} - Q(s_k) \right) \leq M  (1-y^{i}), \forall \,i \in \mathcal W, \label{tt}\\
&  \sum_{k \in \mathcal S_p}^{} \left( \sum_{w  \in \mathcal W}^{} x^i_{wk} - Q(s_k) \right) \geq M  (y^{i}-1), \forall \,i\in \mathcal W, \label{bb}\\
&\sum_{w  \in \mathcal W}^{} \sum_{k \in \mathcal S_p}^{} x^i_{wk}  \leq M  y^{i}, \forall \,i\in \mathcal W, \label{cc}\\
&\sum_{w  \in \mathcal W}^{} \sum_{k \in \mathcal S_p}^{} x^i_{wk}  \geq -M y^{i}, \forall \,i\in \mathcal W, \label{dd}
\end{align}
\label{7}
\end{subequations}
Constraints \eqref{tt}, \eqref{bb}, \eqref{cc}, and \eqref{dd} are the result of the big-M method to guarantee that the team leader recruits a team with workers having the required skills. The term $M$ represents the upper bound of $\sum_{w  \in \mathcal W}^{} \sum_{k \in \mathcal S_p}^{} x^i_{wk}$. Notice that the constraints presented in~\eqref{6} and~\eqref{7} have analogous goal but they are adapted for each strategy.

$\bullet$ \textbf{Leader Participation Constraints:}
The following constraint ensures that the chosen leader contributes with a required skill defined by the project.
\begin{equation}
\sum_{k \in \mathcal S_p}^{}x^i_{ik} = 1, \forall \, i \in \mathcal W.\label{8}
\end{equation}

$\bullet$ \textbf{Leader Uniqueness Constraints:}
To guarantee the uniqueness of the leader, we include the following constraint:
\begin{equation}
\sum_{i  \in \mathcal W}y^{i} = 1 \,.\label{9}
\end{equation}

\subsection{Optimization Problem}
The objective of this paper is to hire the most suitable team to complete a CMCS project $p$. To this end, we introduce a general team formation optimization problem for both approaches, and we define it as follows:
\begin{align}
\text{(P):} & \underset{ \underset{ v_{ww'} \in \{0,1\}}{x^i_{wk} \in \{0,1\}}}{\text{ maximize }}
\sum_{i \in \mathcal I } \sum_{w \in \mathcal W}^{}  \sum_{k \in \mathcal S_p}^{} x^i_{wk} \bigg [ \eta_1 \frac{ \hat{S}^i_{w,k}}{\bar{S}}  - \eta_2 \frac{  U^{i}_j }{\bar{U}}
-  \eta_3 \frac{  C_{w,k}}{\bar{C}}  \bigg  ] \notag \\ 
&\hspace{2cm} + \frac{\eta_4}{|\mathcal S_p|-1} \sum_{j \in \mathcal W}^{}  \sum_{w'  \in \mathcal W}^{} v_{ww'}  \frac{R_{ww'}}{\bar{R}},  \notag \\ 
&\text{subject to:} \notag  
\end{align}
\begin{subequations}
\begin{align}
&\sum_{ i \in \mathcal I } \sum_{k \in \mathcal S_p}^{} ( x^i_{wk}+x^i_{w'k})  \geq  2 \times v_{ww'}, \, \forall \, (w,w') \in \mathcal W \times  \mathcal W, \label{ee}\\
&v_{ww'}  \geq \sum_{ i \in \mathcal I } \sum_{k \in \mathcal S_p}^{} ( x^i_{wk}+x^i_{w'k}) -1, \, \forall \, (w,w') \in \mathcal W \times \mathcal W, \label{rr} \\
&(\text{3})-\eqref{5},  \notag\\
&     \begin{cases}
          \eqref{6} &\,\text{if $i=0$,}\\
      \eqref{7}-\eqref{9} &\,\text{if $i\neq 0$.}
     \end{cases} \notag
\end{align}
\end{subequations}

The value of $v_{ww'}$ is computed using a product of the decision variables $x^i_{jk}$. Therefore, we use the standard linearization technique and replace $v_{ww'}$ given in (2) with the constraints~\eqref{ee} and~\eqref{rr}.

The optimization problem in (P) is formulated as an ILP and the solution can be optimally obtained using off-the-shelf software such as Gurobi or CPLEX integrating the branch and bound algorithm or simplex method. Also, note that there are cases where the problem is infeasible, for example when the number of workers $|\mathcal W|$ is less than the number of skills $|\mathcal S_p|$. However, in CMCS platforms, this case is unlikely to occur since, by definition, in large-scale IoT systems, the value of $|\mathcal W| \gg  |\mathcal S_p|$.

Since the ILP solution is proven to be computationally expensive to obtain, we introduce, in the following section, a low-complexity heuristic approach that uses graph embedding techniques and GA to ensure an effective team recruitment.

\section{Proposed Team Recruitment Algorithm}
\label{sec4}
In this section, we present the architecture of the proposed CMCS team recruitment algorithm composed of five components/phases as depicted in Fig.~\ref{highlevelchart}. The algorithm starts with a data cleaning and preparation phase that generates a digestible format to be fed to the GNN-embedder. The latter converts the workers' graph into numerical feature vectors representing the social relations and attributes of each worker. Afterwards, a clustering algorithm is executed to determine communities of similar workers and hence, reduce the search space for the following recruitment phase. Finally, a GA-based algorithm is run to determine a team of skilled and socially connected workers from the obtained set of candidate workers, i.e., maximizing the objective function of (P).

\subsection{Phase I: Data Cleaning and Preparation}
Initially, we assume that the CMCS platform possesses a database that contains the list of the IoT users registered to the system along with their recorded attributes that we have discussed earlier in Section~\ref{Sec2}. Moreover, the platform has a partial knowledge about their social relations with their peers. This knowledge depends on the offline data that are disclosed by the users or based on their experience and activity in the platform. In order to submit a task to the platform, the request needs to submit all the information required so that the platform can determine the suitable team. Before starting the recruitment phase, we propose to perform a data cleansing process on the workers' SN dataset. This phase ensures that all the unknown or missing relationships are found and eliminated. Also, we perform the conversion of the graph $\mathcal G(\mathcal W,\mathcal E)$ to a fully connected graph while respecting the defined model earlier. In other words, we associate the value $\hat{R}_{ww'}$ between each pair of workers $(w,w')$.

\subsection{Phase II: Network Embedding}
\textcolor{black}{Afterwards, we proceed with a network embedding phase. Network embedding is a technique that converts the 3D structure of the graph and embeds it into a lower dimensional continuous latent space (i.e., numerical representation of the graph vertices)~\cite{8294302,8047276}. Since vector operations are often simpler and faster than the equivalent graph operations, graph embedding leverages the data-science tool sets by learning a mapping from a 3D network to a 2D compressed representation space while preserving relevant network properties.}

\begin{figure}
    \centering
    \includegraphics[width=7.5cm]{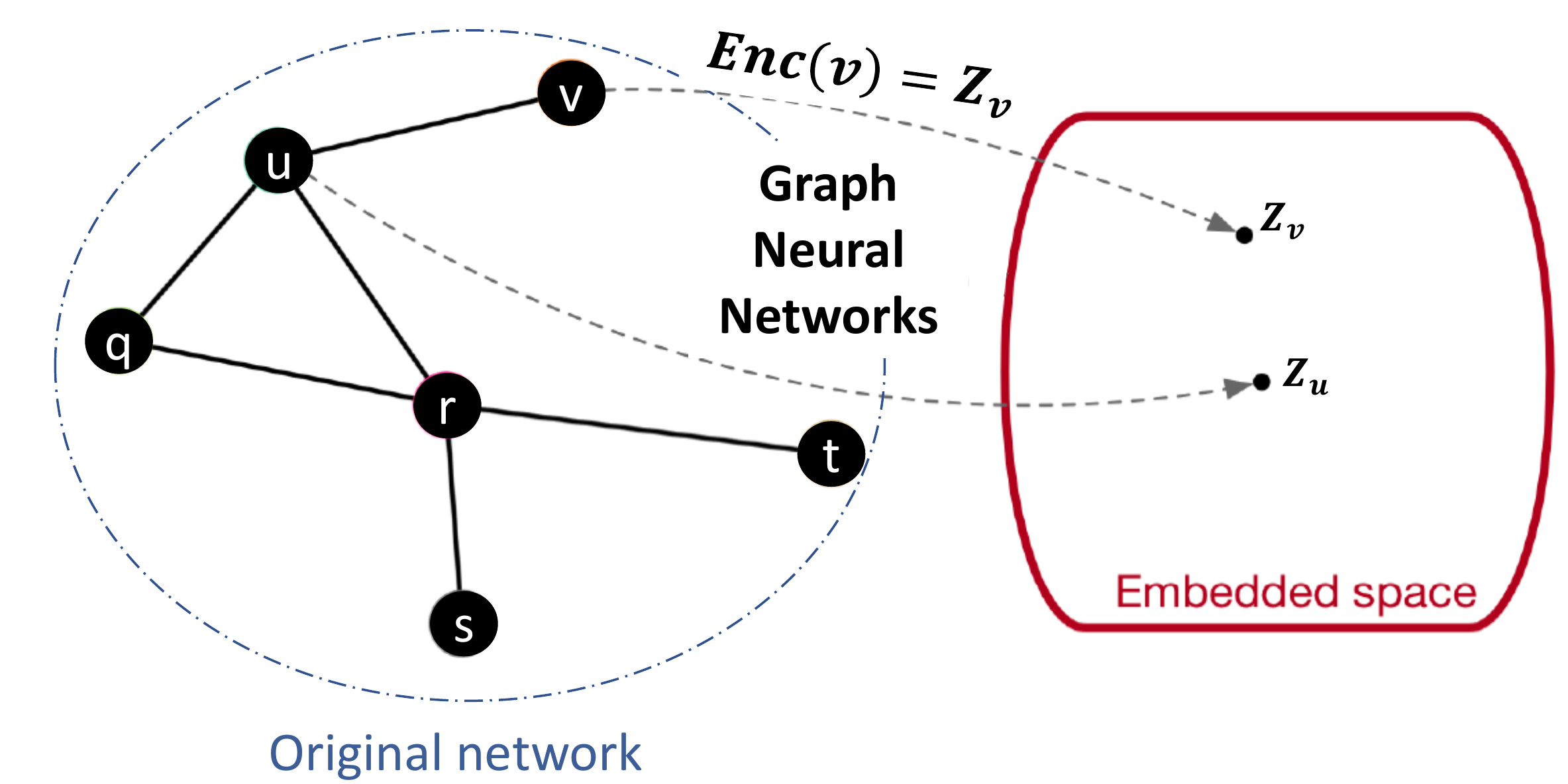}
    \caption{\textcolor{black}{Visualization of the embedding process where the original 3-D network is transformed into an embedding space with 2-D vectors by the means of an encoding function $ENC()$.}}
   \label{fig:embeddingexplain}
\end{figure}

\textcolor{black}{As recent work has shown, there are various ways to go about embedding graphs, each with a different level of granularity~\cite{zhu2020cagnn,hamdi2020flexgrid2vec}. The embedding can be performed on the on the node level~\cite{grover2016node2vec} or on the entire graph~\cite{narayanan2017graph2vec}. In node embedding, each vertex (node) is encoded with its vector representation. This approach is used if the objective is to perform visualization or prediction on the vertex level, e.g., visualization of vertices in the 2D plane or prediction of new connections based on vertex similarities. By definition, the nodes close to each other in the graph, relying on specific measure, must be close to each other in the embedding space, using that same me sure. Deepwalk~\cite{DBLP:journals/corr/PerozziAS14} by Perozzi et al. was the first approach that tackled the node embedding group. It uses random walks to produce embeddings. However, this technique has several limitations, including the disregard of the neighborhood of the embedded node. Node2vec~\cite{grover2016node2vec} approach by Grover et al. surpasses this issue by proposing a different walking technique between nodes. It has parameters $P$, and $Q$. Parameter $Q$ defines how probable it is that the random walk would discover the undiscovered part of the graph, while parameter $P$ defines how probable it is that the random walk would return to the previous node. Graph embedding, on the other hand, is a technique that transforms the whole graph is into a single vector~\cite{9382981}. This approach is used to perform predictions at the graph level or compare and visualize entire graphs, e.g., comparing chemical structures. Methods that tackled this group are Graph2vec~\cite{narayanan2017graph2vec} which are based on the idea of the doc2vec approach and rely on node2vec.}

\textcolor{black}{It is worth to note that there is another important distinction between the embedding techniques. There is a transductive embedding which is a low-dimension vector representation derived for each node making it impossible to find the vector representation for a new node. In data science terms, the algorithm cannot perform prediction based on unknown data. Transductive methods are principally based on matrix factorization techniques and random walks. This approach is simple and is able to generate embeddings with fewer nodes. The inductive embedding on the other hand is more consistent with the common training/prediction model that we are used to find in most machine learning techniques. It requires labeled data in order to train the target function.}

\textcolor{black}{
In our proposed approach, we are interested in extracting information from the workers' network structure and from each of their devices in a simple fast way. Therefore, we focus on the transductive node embedding.} There are two groups of embedding in the social network graph: 1) the first one is a featureless node embedder using single layer GNNs, \textit{aka}, shallow embedders and 2) the second one is an attributed graph embedder using unsupervised multi-layer GNNs. In the first group, the embedding captures only the social network graph topology, vertex-to-vertex relationship, and sometimes other relevant information about graphs, subgraphs, and vertices. The second group, however, is also able to capture the nodes' characteristics and features while including them in the resultant output embedding vectors. The key difference between these two groups lies in the definition of the encoding function mapping the social network graph to the embedding space. As Fig.~\ref{fig:embeddingexplain} illustrates, this function, referred to as $ENC()$, is responsible from outputting the node $u$'s embedding vector value $Z_u$ from the original network. Depending on which embedding group (i.e., single layer GNNs or unsupervised multi-layer GNNs), the function $ENC()$ must be carefully defined in order for the resultant vector value $Z_u$ to include either the social network vertex-to-vertex relationship or both vertex-to-vertex relationship and the devices' features.

\textcolor{black}{
For the first group, this mapping function represents a simple encoder that captures only the workers' social relationships, which can be written as follows~\cite{hamilton2018representation}:
\begin{align}
    ENC(u)= \mathbf{z}_{u} = Z \times e_u,
\end{align}
where $Z$ is the output matrix with dimension $d \times |\mathcal V|$ containing the embedding values of the nodes $V$ and $e_u$ is an indicator vector that has all values set to zero except in one column, set to one, indicating the presence state of node $u$.}

\begin{figure}
    \centering
    \includegraphics[width=9cm]{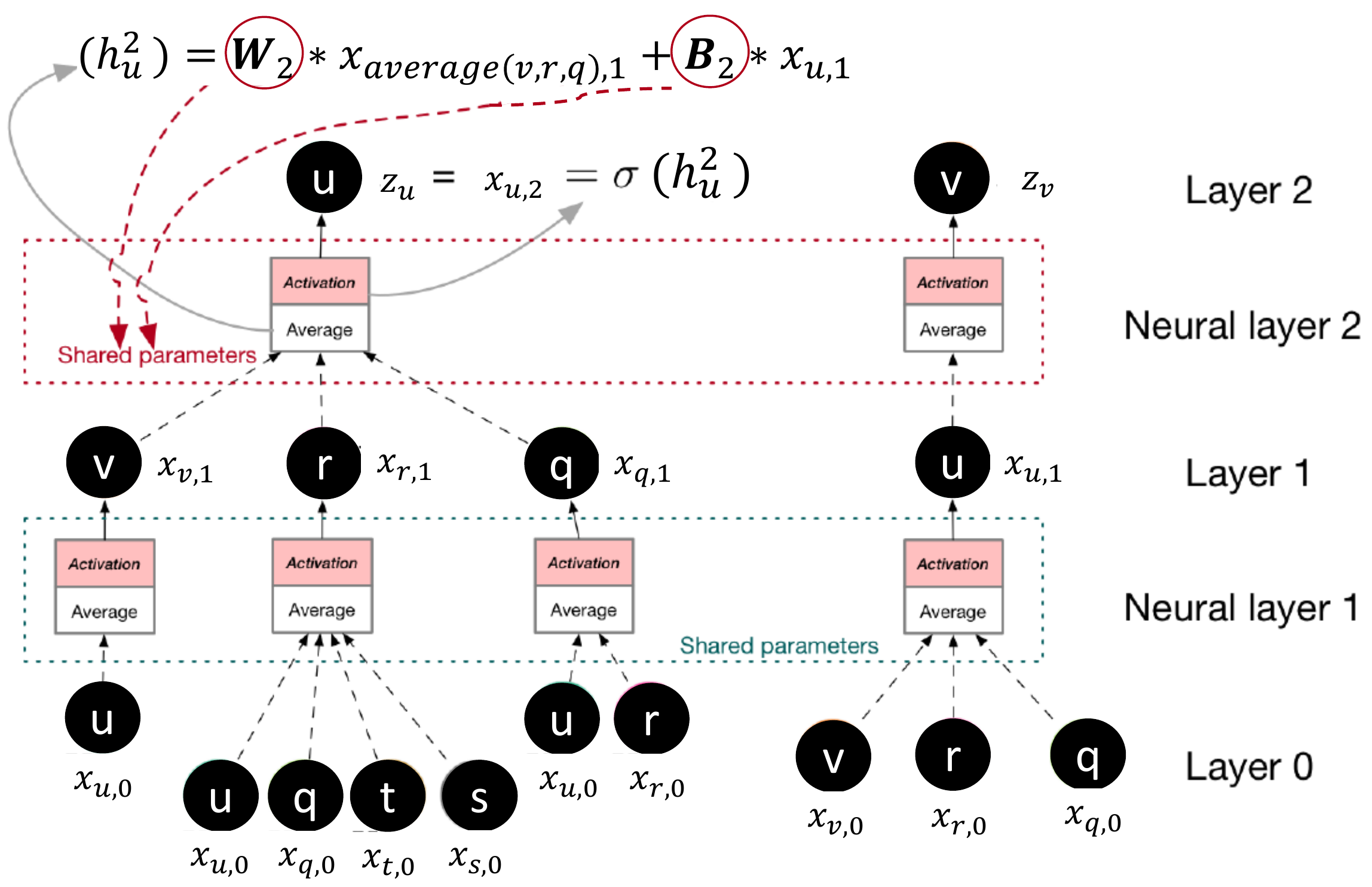}
    \caption{\textcolor{black}{Illustration of the graph neural network embedding two nodes $u$ and $v$ using their attribute vector $x_{u,0}$ and $x_{v,0}$. }}
\label{fig:gnnillustrations}
\end{figure}

\textcolor{black}{
For the second group, and because of its complexity and need to capture the nodes' attributes, the encoder is in the form of unsupervised multi-layer neural networks~\cite{zhou2021graph}. The encoder no longer only takes the node's relations as input but also its attributes. As Fig.~\ref{fig:gnnillustrations} shows, the encoder is a computational graph with multiple encoding layers. The first layer (i.e., layer 0) takes as input vector $X_{u,0}$, which also involves the attributes of the node $u$. The encoding for node $u$ is accomplished using $X_{u,0}$ and all the directly connected nodes in a set $N(u)$, which, in the example of Fig.~\ref{fig:gnnillustrations}, are nodes $q$, $r$, and $v$ for node $u$.  The final layer, in  Fig.~\ref{fig:gnnillustrations}, Layer 2, captures the embedding of the nodes.  The neural layers corresponding to layer $k$ for a node $u$ are defined as $h_{u}^{k}$ and can be written as follows:
\begin{align}
&\mathbf{h}_{u}^{k}=&\sigma\left(\mathbf{W}_{k} \sum_{a \in N(u)} \frac{\mathbf{h}_{a}^{k-1}}{|N(u)|}+\mathbf{B}_{k}\notag \mathbf{h}_{u}^{k-1}\right), \\&& \forall k \in\{1, \cdots, K\} 
\end{align}
The initial 0-th layer embedding $\mathbf{h}_{u}^{0}$ are equal to the node features $\mathbf{x}_{u}$ and the optimized embedding $\mathbf{z}_{u}$ are equal to the final layer embedding $\mathbf{h}_{u}^{K}$. The $\sigma$ function represents the non-linearity (e.g., relu) and  $\mathbf{W}_{k}$ and $\mathbf{B}_{k}$ represent the trainable weight matrices that will be adjusted with the loss function. The term $\sum_{a \in N(u)} \frac{\mathbf{h}_{a}^{k-1}}{|N(u)|}$ represents the average of neighbors's previous layer embedding. A more simplified vector format is written as follows:
\begin{equation}
\begin{array}{l}
\mathbf{H}^{(l+1)}=\sigma\left(\mathbf{H}^{(l)} \mathbf{W}^{(l)}+\tilde{\mathbf{A}} \mathbf{H}^{(l)} \mathbf{B}^{(l)}\right),
\end{array}
\end{equation}
where $\tilde{\mathbf{A}}=\mathbf{D}^{-\frac{1}{2}} \mathbf{A} \mathbf{D}^{-\frac{1}{2}}$ and $\mathbf{H}^{(k)}=\left[\mathbf{h}_{1}^{(k)^{T}}, \ldots,\mathbf{h}_{m}^{(k)^{T}}\right]^{T}$, where $\mathbf{D}$ is the  diagonal matrix, $\mathbf{A}$ is the social relationship adjacency matrix and $m$ represents the number of embeddings in layer $k$.}

\textcolor{black}{After defining the appropriate encoders for the single layer GNN and unsupervised multi-layer GNN, we need to define two similarity functions~\cite{zhou2021graph} that specify how the nodes characteristics in the vector space map to nodes' characteristics in the original network. The first similarity function, denoted by $S$,  compares the nodes in the original graph, while the second function compares the embedding vectors. The similarity between tow nodes $u$ and $v$ in the embedding space is simple and can be defined as $\mathbf{z}_{v}^{\top} \mathbf{z}_{u}$, the dot product between the vectors. The goal is to find a function $S$ where the embedding $z_u$ and $z_v$ of the nodes $u$ and $v$ can be optimized in a way such that:
\begin{equation}
S(u, v) \approx \mathbf{z}_{v}^{\top} \mathbf{z}_{u}.
\label{simembedding}
\end{equation}}

\begin{figure*}[t]
\begin{minipage}[h]{1\linewidth}
    \centering
  \frame{  \includegraphics[width=18cm]{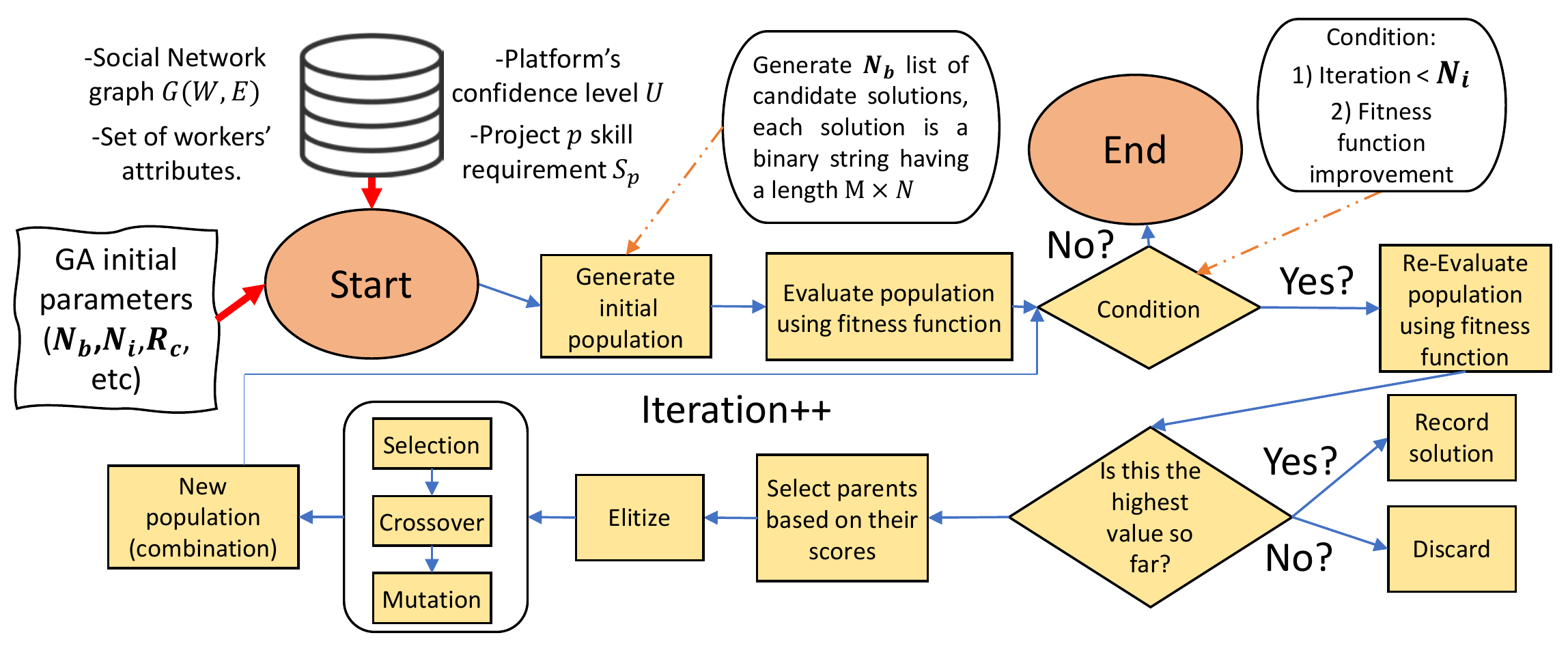}}
    \caption{Flowchart illustrating the worker selection mechanism (Phase V) of the proposed CMCS recruitment algorithm.}
    \label{process}
\end{minipage}
\end{figure*}
\textcolor{black}{
In the literature, there have been some studies~\cite{Wu_2021,shao2021learning} focusing on multiple approaches using different similarity functions. Some functions include a basic adjacency matrix, multi-hop network similarity matrix, or a random walk similarity~\cite{9039675,9339909,9306765}. For the shallow encoding, we rely on the random walk technique. The nodes similarity in the social network graph is computed based on the probability that two nodes $u$ and $v$ co-occur on a random walk over the network.
Moreover, the probability of visiting node $v$ on a random walk starting from node $u$ using some random walk strategy $R$ describes the similarity between the two nodes. We choose this particle similarity function because it incorporates both local and higher-order neighborhood information and does not need to consider all node pairs when training; it only needs to consider pairs that co-occur on random walks.
The difference between this similarity, also known as Loss, is the function that will be optimized, and it is written as follows:
\begin{equation}
\mathcal{L}=\sum_{u \in V} \sum_{v \in N_{R}(u)}-\log \left(P\left(v \mid \mathbf{z}_{u}\right)\right),
\end{equation}
where $N_R(u)$ describes the multiset of nodes visited on random walks starting from node $u$ and  $P\left(v \mid \mathbf{z}_{u}\right)$ represent the likelihood of random walk co-occurrences between the node $v$ and embedding of node $u$ computed using Softmax as follows:
\begin{equation}
P\left(v \mid \mathbf{z}_{u}\right)=\frac{\exp \left(\mathbf{z}_{u}^{\top} \mathbf{z}_{v}\right)}{\sum_{n \in V} \exp \left(\mathbf{z}_{u}^{\top} \mathbf{z}_{n}\right)}.
\end{equation}
This term represents the predicted probability of $u$ and $v$ co-occuring on random walk.
The overall loss function that needed to be minimized is: 
\begin{equation}
\mathcal{L}=\sum_{u \in V} \sum_{v \in N_{R}(u)}-\log \left(\frac{\exp \left(\mathbf{z}_{u}^{\top} \mathbf{z}_{v}\right)}{\sum_{n \in V} \exp \left(\mathbf{z}_{u}^{\top} \mathbf{z}_{n}\right)}\right),
\label{lossfunction}
\end{equation}
where the first sum is over all nodes $u$ in the social network, the second sum is over all nodes $v$ seen on random walks starting from $u$ using a strategy $R$. Optimizing random walk embeddings means finding  $z_u$  that minimizes $\mathcal L$. The unsupervised multi-layer GNN approach takes advantage of the same loss functions with a stochastic gradient descendent. In this paper, we utilize the same loss function defined in \eqref{lossfunction} for the single-layer GNN.}

\textcolor{black}{
In our proposed approach, we utilize both the featureless node embedders and the attributed graph embedders. As shown in Fig.~\ref{highlevelchart}, for the featureless node embedder, we have only one input (i.e., $input_1$ only) and we employ the Graph Embedding with Self Clustering (GEMSEC)~\cite{gemsec}).  For the attributed graph embedder,  we have two inputs (i.e., $input_1$ and $input_2$) and we employ the Attributed Social Network Embedding (ASNE)~\cite{ASNE}). The GEMSEC provides a limited knowledge about the workers. It only embeds their social relationships and allows later clustering workers accordingly. However, the ASNE embedder provides a better knowledge about the whole system by including the workers' social relations as well as their skills so that the obtained clusters will regroup workers that are socially connected and share similar expertise. This comes at the cost of a higher implementation complexity as it requires large and diverse dataset.}

\textcolor{black}{
The output of this phase is $|\mathcal W|$ feature vectors that are forwarded to the clustering phase. The embedding is done in order to turn the network graph into a set of numerical feature vectors that can be handled by machine learning models.}

\subsection{Phase III: Clustering}
Once the vector representation for each node in the SN graph is determined, we proceed with a dimensional reduction algorithm (e.g., T-SNE or PCA) in order to reduce the number of variables of the data by extracting most important ones from a large pool. This phase alleviates the clustering problem, speeds up the computation process, and presents a visualization method in the 2-dimensional space for the clusters. After that, an unsupervised machine learning technique can be utilized to group the workers with common features and attributes into clusters or communities. If the GEMSEC is used in the embedding phase, the workers that are socially connected with be grouped in the same clusters. These workers may have different skills. We define this method as the "Edge-only Embedding". However, if the ASNE is employed, then the workers sharing strong social relations and similar skills will be labeled in a common cluster. We identify this method as the "Edge-Attribute Embedding".

There are several possible machine learning approaches for clustering. The most known ones are Affinity Propagation, Agglomerative Clustering, DBSCAN, $K$-Means, OPTICS, Spectral Clustering, and Mixture of Gaussians~\cite{7976553}. As shown in Fig.~\ref{highlevelchart}, the inputs to this phase are feature vectors that the clustering algorithm will be performed on. The outputs are the resulting communities, which will be used as a small search space for the GA-based recruitment algorithm.

\subsection{Phase IV: Data Post-processing}
Given the obtained groups of similar workers, a cluster selection algorithm is executed during a data post-processing phase to output a set of candidate workers by determining the appropriate clusters that should be forwarded to the recruitment phase. 
The objective of the post-processing phase is to determine the most appropriate cluster, or clusters,  that the recruitment algorithm of the following phase should focus on.

For the Edge-only embedding, our solution is to assign to each cluster a "score" that reflects the degree of appropriateness. This score is computed based on the average efficiency of workers (equation~\eqref{1}) belonging to each cluster. It takes into account the skills that are required by the project's cluster $\mathcal S_p$, average uncertainty level of the recruiter $i$, average requested reward of the workers, and average SN strength. Hence, the most appropriate cluster is chosen as follows:
\begin{equation}
 c^* = \argmax_{c=1,\dots, \bar{C}} \sum_{w \in \mathcal C_c}\sum_{k \in \mathcal S_p } \frac{E^i_{w,k}}{|\mathcal C_c|}, \quad \forall i,
\end{equation}
where $\mathcal C_c$ is the cluster $c$ obtained from the clustering phase and $\bar{C}$ is the total number of clusters. Next, the best cluster~$c^*$ is selected for the next phase.

For the Edge-Attribute embedding, the clusters are grouping workers having similar skills. Hence, the data pre-processing phase aims to determine the most suitable $|\mathcal S_p|$, i.e., each cluster representing a particular skill. To this end, it assigns to each of the clusters a skill score, which is computed as the average skill level of all members of the cluster as known by the recruiter $i$. A skill $k^*_c$ is assigned to cluster $c$ as the most representative one using the following formula:
\begin{equation}
    k^*_c = \argmax_{k \in \mathcal S_p}\sum_{w \in \mathcal C_c} \frac{\hat{S}^i_{w,k}}{|\mathcal C_c|}, \quad \forall i,\quad \forall c=1,\dots, \bar{C},
\end{equation}
In this way, the recruitment algorithm of the following phase will consider only the obtained $|\mathcal S_p|$ clusters and, from which, will select at most one worker.

\subsection{Phase V: Worker Selection}

For the worker selection phase, we propose to employ a low-complexity approach based on GA~\cite{7563421} that will be applied on the list of workers obtained from the data post-processing phase. The GA algorithm solves the CMCS recruitment problem formulated in (P) but on a much smaller search space compared to the ILP. Indeed, this algorithm is proposed as an alternative to the NP-hard optimal ILP approach because of its low overhead and running time. 
GAs simulate the process of natural selection which means those species who can adapt to changes in their environment are able to survive and reproduce and go to next generation.  
In GA, each generation consists of a population of individuals and each individual represents a point in the search space and a possible solution. 

\textcolor{black}{Once we determine the most appropriate shortlisted set of candidate workers after the post-processing phase, the GA-based worker selection algorithm starts the process of matching workers and forming the appropriate team for the project of interest. The GA's solving workflow is illustrated in Fig.~\ref{process} and can be briefly summarized as follows: At first, an initial generation of genes is created. This generation can be a string of digit values where each digit reflects a certain status. This generation evolves using: i) the selection operator which is mainly giving preference to the individuals with good fitness scores and allows them to pass their genes to the successive generations,  ii) the crossover operator which represents mating between individuals, and hence, two individuals are selected using selection operator and crossover sites are chosen randomly, and finally iii) The mutation operator which inserts random genes in offspring to maintain the diversity in population to avoid the premature convergence. This process is repeated until a pre-defined condition is met. As shown in Fig.~\ref{process}, this condition can be, for example, the number of iterations reaching a threshold value $nI$  or when there is no improvement in the fitness function. In the case of looping a local optimum, we apply a restart operation which restarts the whole process with new initial generation.  We choose to perform elitism after each generation to ensure that the best individual remains in the next generation. In the case of platform-based scheme, the genes population represents the series of workers' skill attributes. In fact, each worker is represented by an $S$-digit binary values where $S$ represents the possible skills in the platform. For example, if $|S|=4$, and $|\mathcal W|=3$ then $"100000100001"$ represents the genes series of three workers, where the first worker contributes with the first listed skill in the platform, the second worker contributes with the third skill in the platform, and the third worker contributes with the last listed skill in the platform.}

\textcolor{black}{We should note that when generating the initial population, the default GA generates random strings. However, in our case, we tune this process and setup a specific generation technique that is able to enhance the performances of the GA. In fact, since all the feasible solutions are strings with length $N\times M$ having only $|S_p|$ set of ones, we opt to modify the random generation to produce strings with length $N\times M$ containing  $|S_p|$ set of ones and ($N\times M-|S_p|$) set of zeros. In this way, the algorithm initially starts with a feasible solution and the search space is narrowed down, consequently boosting the GA performances and convergence speed.  However, this comes at the expense of increasing the likelihood of getting stuck in local maxima/minima. In order to overcome this and avoid being trapped a local optimum, we have introduced the restarting technique to completely re-initiate the GA algorithm each time the algorithm detects a premature convergence.} 

\textcolor{black}{Also, in the case of multiple project assignments, the string of genes is no longer binary but rather integer with values taking place between $0$ and $|\mathcal P|$. In the case of the leader-based scheme, we could add a series of bits with length $|W|+1$ at the beginning of the gene string stating the recruiter. We also should reduce for this specific string the frequency of mutation and crossover since any change on this particular series of bits is sensible as it changes the recruiter.}

\section{Experiments and Evaluation}\label{sec5}
In this section, we investigate the performances of the proposed CMCS recruitment approaches. At first, we start by presenting the used dataset and environmental setup. Then, we conduct experiments to show the differences between the leader-based and platform-based recruitment strategies using the ILP method. Next, we analyze the behavior  of the proposed GNN approach by investigating the embedding and clustering performances to, finally, close the loop by comparing the performances of the GNN recruitment method with the ones of the ILP. \textcolor{black}{We should note that all the simulations to evaluate the performances of the proposed GNN-based approach were done on the entire large-scale dataset presented in the following section. However, whenever the ILP was involved (either when analyzing the performances of the two recruitment schemes or when validating the performances of the proposed GNN-based approach), we had to shrink the huge initial pool of workers to a small-scale dataset because of its high computational complexity.}

\begin{figure}[t]
\vspace{0.25cm}
\frame{\includegraphics[width=8.5cm]{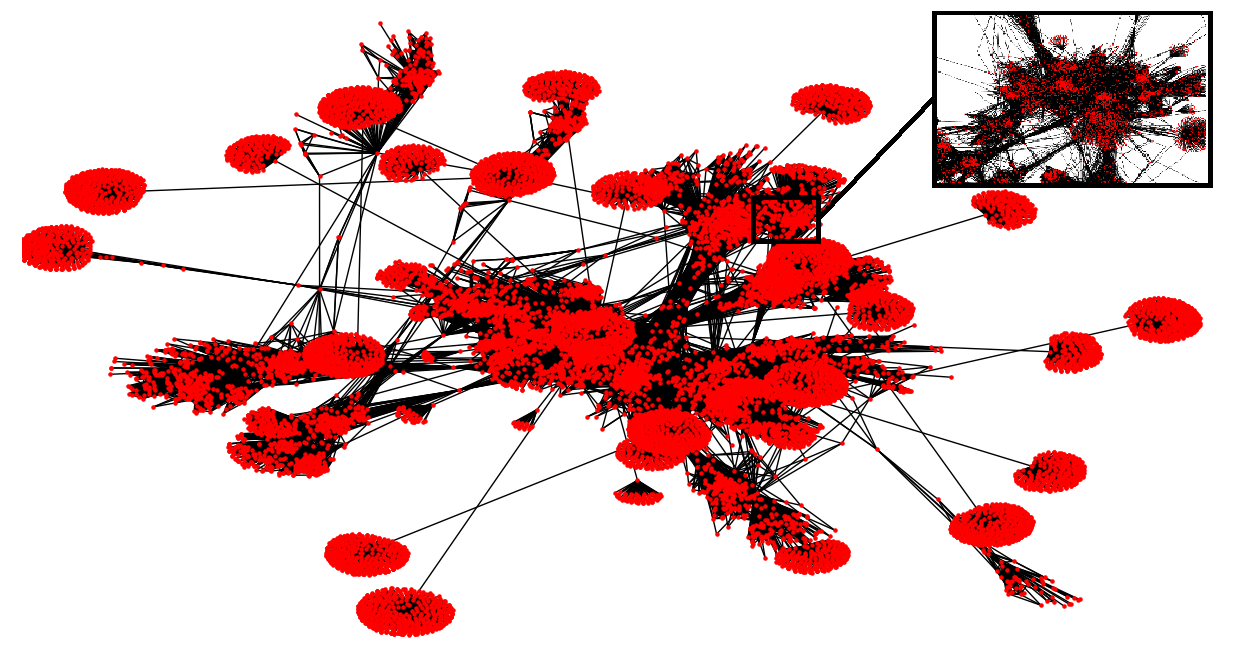}}
    \caption{The SN graph extracted from the real-world Facebook dataset.}
    \label{facebooknetwork}
\end{figure}
\begin{figure}[t!]
        \includegraphics[width=8.5cm]{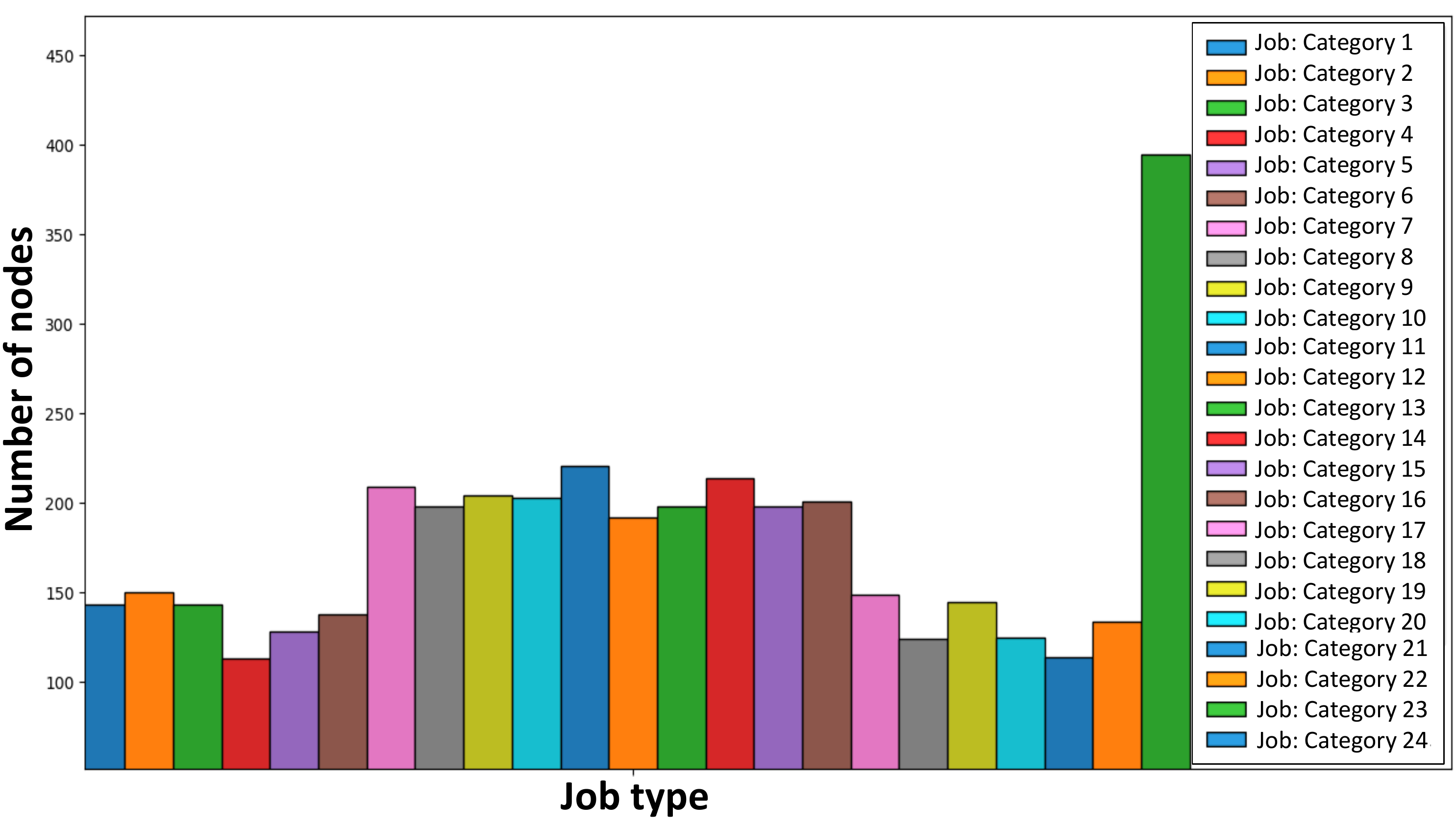}
    \caption{Histogram illustrating the Job type distribution in the Dataset.}
    \label{jobtype}
\end{figure}

\begin{figure}[!t]
        \includegraphics[width=8.5cm]{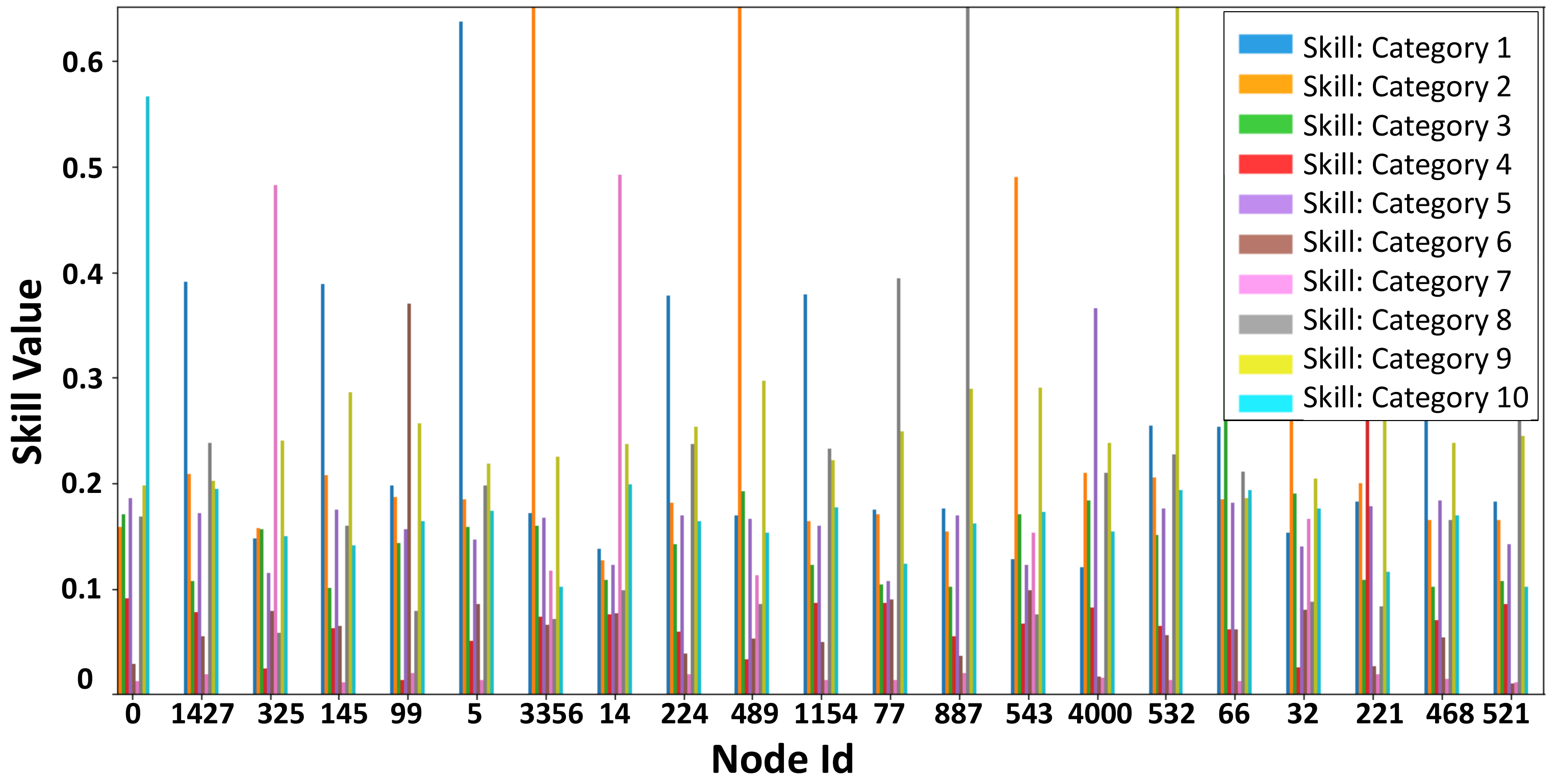}
    \caption{Samples from the dataset showing the skill values for certain workers. A work can have multiple skills at the same time but with different levels.}
    \label{sampleskill}
\end{figure}
\begin{figure*}[!ht]
  \centering
  \begin{subfigure}
    \centering
    %Average Community Size
    \includegraphics[width=4.25cm,height=3.5cm]{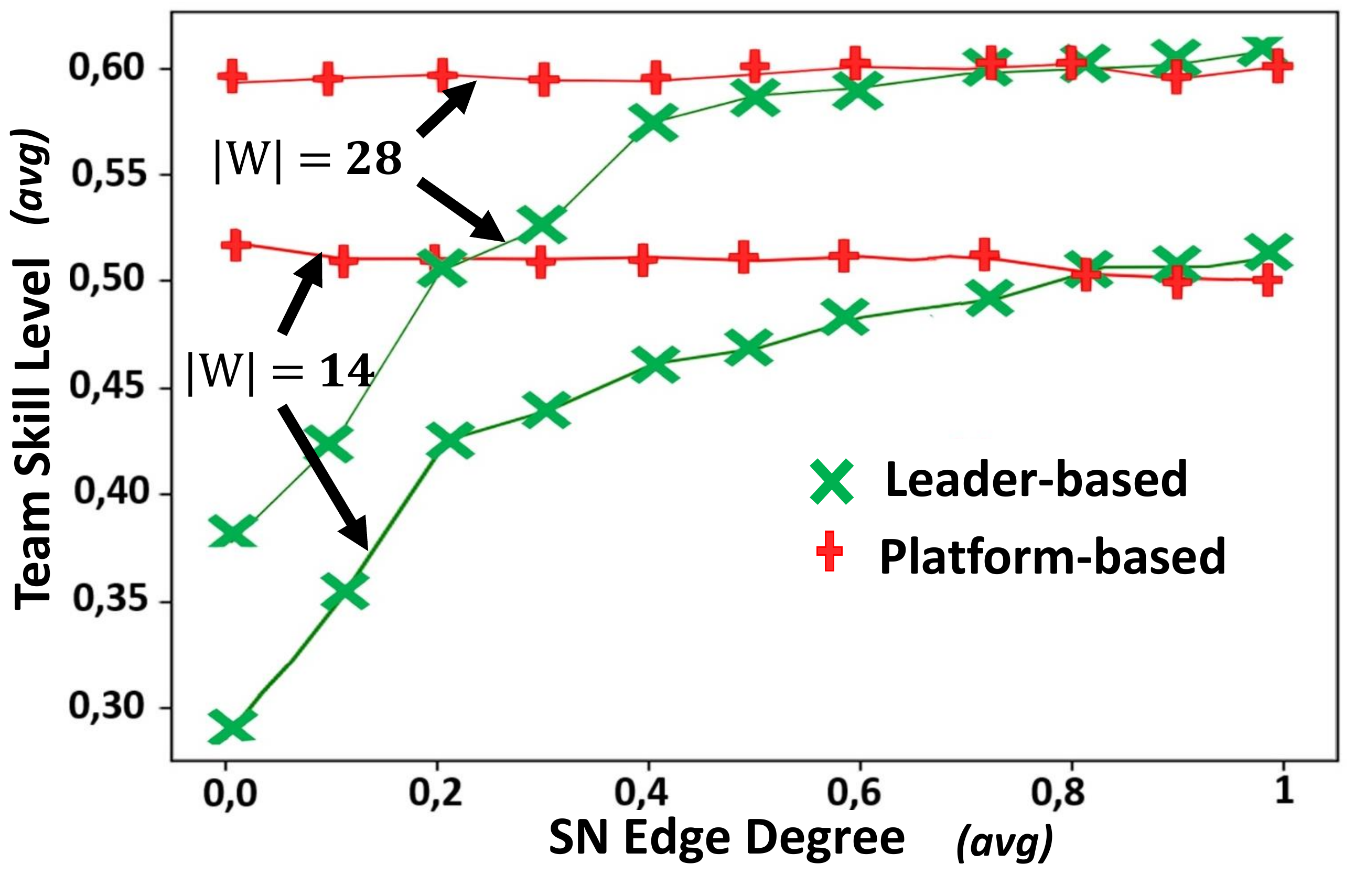}
  \end{subfigure}%
 ~ \begin{subfigure}
    \centering
    % Time to process
    \includegraphics[width=4.25cm,height=3.5cm]{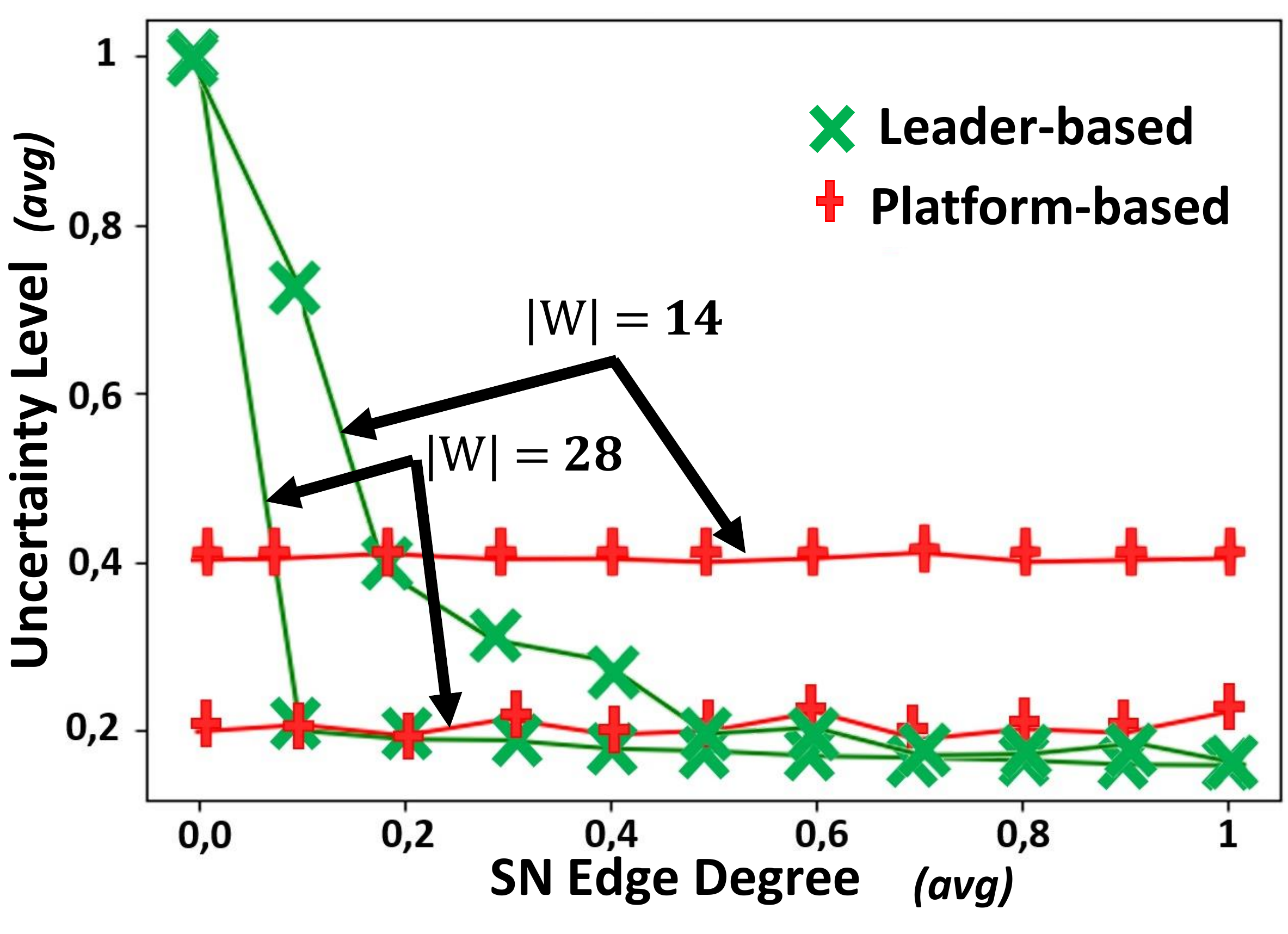}
  \end{subfigure}
  ~  \begin{subfigure}
    \centering
    %Average Community Size
    \includegraphics[width=4.25cm,height=3.5cm]{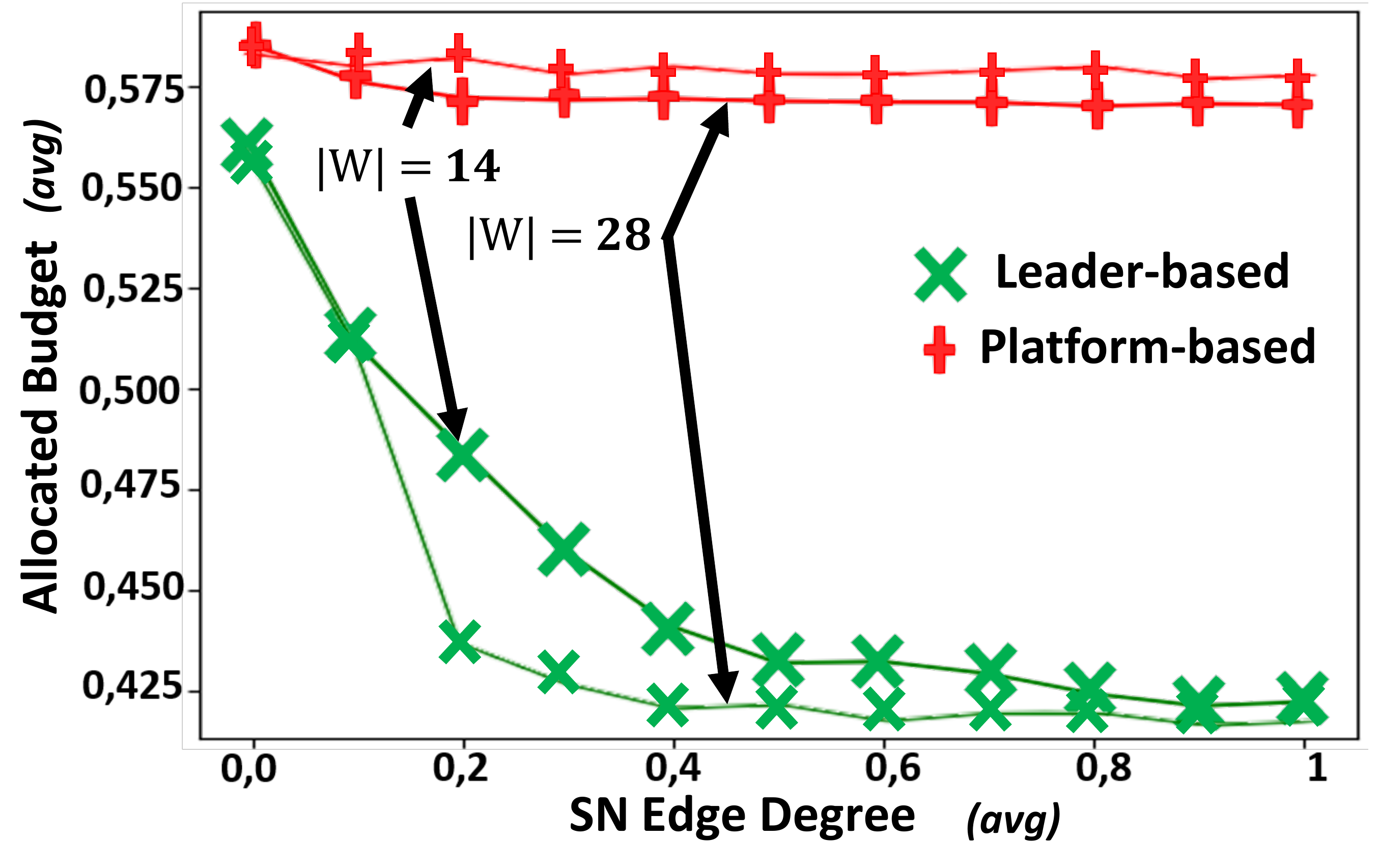}
  \end{subfigure}%
  ~\begin{subfigure}
    \centering
    % Time to process
    \includegraphics[width=4.25cm,height=3.5cm]{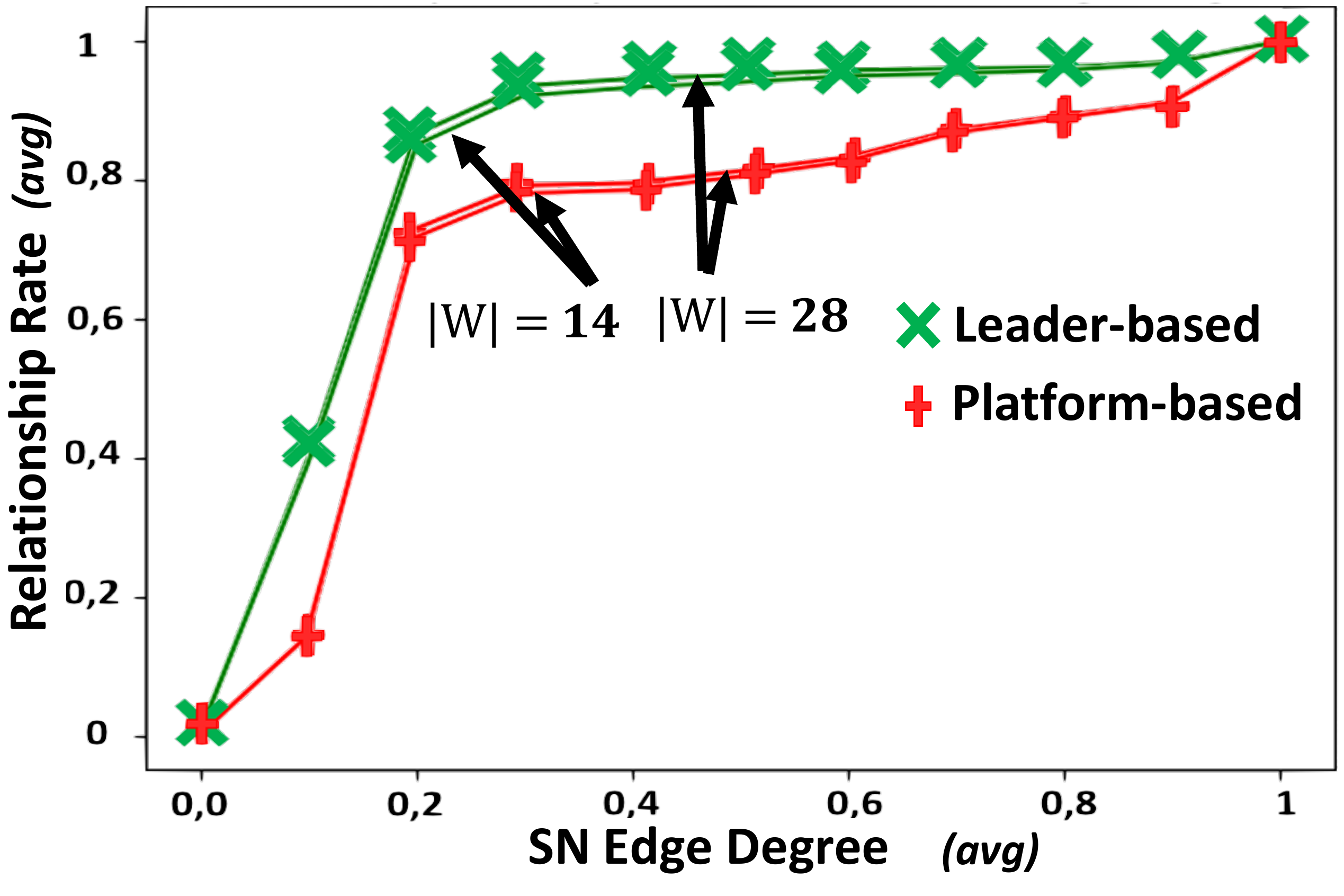}
  \end{subfigure}
 
    \caption{Efficiency criteria  vs. SN edge degree for the leader-based and the platform-based recruitment strategies with $|\mathcal S_p|=5$ for values of $|\mathcal W|$ set to $14$ and $28$. (a) Average skills level, (b) average recruiter's uncertainty level of the recruited team, (c) average cost per worker, and (d) social relationship rate. }
  \label{fig:results} 
\end{figure*}

\subsection{Dataset and Environmental Setup} 
Because there are no available real-world datasets that can be directly used to simulate CMCS applications, we had to fine-tune our own dataset while inspiring from the most suited available ones in order to fairly perform the evaluation.

Our created dataset is partially synthetic and exploits the "ego-Facebook"~\cite{NIPS2012_4532}, a real dataset from Facebook collected from $4039$ survey participants and containing $88234$ edges, to simulate the workers' social network.  The "ego-Facebook" is initially an ego-centered dataset that contains lists of 'circles' (i.e., 'friends lists') from Facebook. This can be seen in Fig.~\ref{facebooknetwork} where the users' circles appear like communities. The dataset originally includes the nodes' edges and their corresponding features (i.e., profile attributes). Examples of these features are the Facebook id, job type, work experience, sign-up date, and birthday. Because of the privacy concerns, all the data has been anonymized and the feature's interpretation has been obscured and replaced with anonymized data. For instance,  if a feature "Area-of-Expertise" has a category "Computer Systems", the "ego-Facebook" dataset would simply contain "Area-of-Expertise=anonymized category 1". A histogram illustrating the job feature with multiple categories is shown in Fig.~\ref{jobtype}.  Since the data has been obscured, we choose to manually label each of the categories such that they are related to CMCS applications (e.g., WorkCategory1 is turned to "doctor", workCategory is turned to "salesman", etc.). To calculate and assign the workers' skills levels and demanded cost, we extract and join some of the features that we find appropriate to CMCS applications such as current job, previous job, work domain, and work start date and join them in a while introducing our own fuzzy-logic values.  Moreover, we use the Facebook users' current field of work (e.g., IT, medical, business, etc.) and their education background (e.g., high school, university, school) along with their availability (e.g., retired, full-time, unemployed) to estimate the degree of skills and demanded cost of each worker. For example, users with a medical field of work and have a full-time availability are more qualified to achieve a project that requires high medical skills than a nurse with high school background and part time availability. Furthermore, if a certain CMCS project demands skilled workers in the medical field and the platform contains profiles of users which have 'doctor, nurse, etc.', it is only natural for a node with a category "doctor" to be more skilled than the other two categories.
For the reward attribute, we introduce some randomness to reflect the human demand noise. We also introduce another randomness where we assign to every node some degrees of skill level in the fields out of its main specialty. This can be seen in Fig.~\ref{sampleskill} where a worker that has a domain of expertise in IT (e.g., category 1) can have little background in the medical field (e.g., category 4), therefore, we choose to attribute a low value of skills for all the workers in their non-domains of expertise. As such, we do for the rewards.

\begin{figure}[t]\vspace{0.05cm}
\centering
    \frame{ \includegraphics[width=8.75cm]{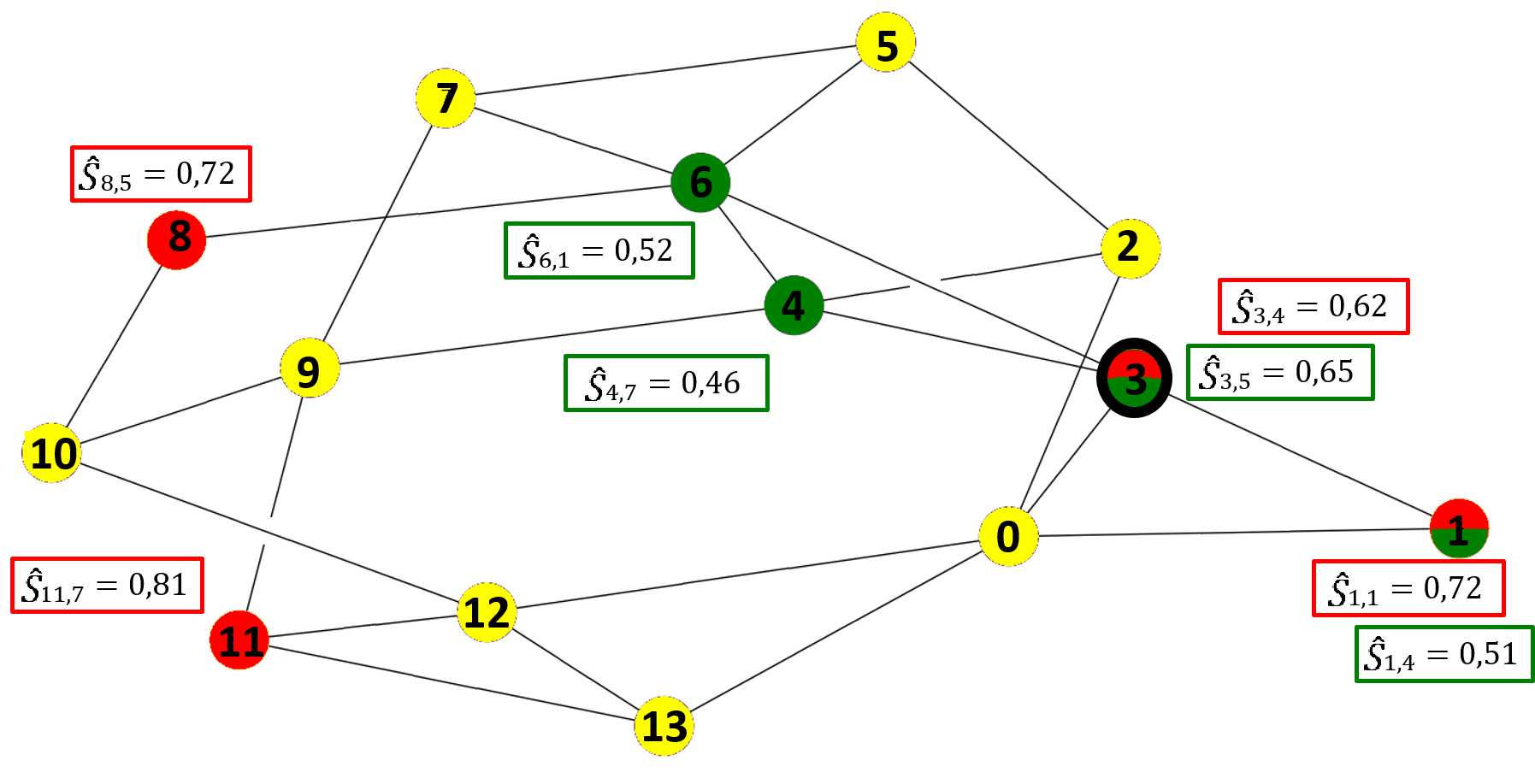}}
\caption{An example of the selected teams using both recruitment strategies for an SN graph where $|\mathcal W|=14$ and $|\mathcal S_p|=4$. The selected team members are green and red for the leader-based and the platform-based approaches, respectively. For the former approach, the team leader is contoured in bold.}
\label{2figures}\vspace{-0.2cm}
\end{figure}

To sum up, our resultant dataset contains $88234$ real world relationships between $4039$ ego-centered Facebook users. We took the Facebook circles, their ego center, and the edges and built the SN relationship graph. Then, we used Dijkstra shortest path algorithm in order to turn the graph into a fully connected one where the edges are inversely proportional to the number of hops between the users. This way, we can quickly estimate the distance between users without having to compute it every time in the algorithm. For each of the users, we estimated their attributes, i.e., their skill and reward, using the real features contained in the Facebook dataset. These two attributes are fuzzy-logic based as discussed in \ref{workermodel} and they are computed using the users' history of job field, experience, education, working period, and availability.

The values of $\omega_i$ in the objective function of (P) are chosen uniformly (i.e., $\eta_i=\frac{1}{4}$, $\forall i$). In our experiments, all algorithms are implemented in a Python 3.6 environment and run on a 32 socket Intel(R) Xeon (R) E5-2698 v3 @2.30GHz CPU with 72G of RAM. To solve the ILP problems, we use the python API of academical CPLEX. For the first embedding approach (i.e., using GEMSEC), we set the number of node embedding dimension to 23, the random walk length to 80, the number of walks from source to 5, the downsampling distortion to 0.75. For the second approach (i.e., using ASNE),  we set the number of feature embedding dimension to 48, the size of the gradient to 64, and the number of training epochs to 10. For the following sections, we refer to $|\mathcal W|$ as the number of workers selected randomly from the pool of $4039$ workers available in the dataset.

For the platform-based approach, the knowledge level about the workers' skills and the workers' relationships are proportionally modeled to their history in the platform (i.e., workers with more history in the platform has lower uncertainty levels). On the other hand, the uncertainty levels of potential leaders $i$ towards workers $w$ for the leader-based recruitment strategy regarding the workers relationships are proportional and increase with the number of hops between the team leader and other workers. For both approaches, the error is modeled as a zero-mean normal distribution $\sim \mathcal{N}(0,\,0.3^2)$.

\subsection{ILP Recruitment: Comparison between Platform-based Strategy and Leader-based Strategy}

In Fig.~\ref{fig:results}, we perform a Monte Carlo simulation where $1,000$ realizations of different parameter settings are generated. We evaluate the average four metrics of the selected teams: skills efficiency Fig.~\ref{fig:results}(a), recruiter confidence Fig.~\ref{fig:results}(b), team cost Fig.~\ref{fig:results}(c), and social relationships Fig.~\ref{fig:results}(d). Due to the high complexity of the ILP approach, we randomly pick a small subset $|\mathcal W|$ from the entire dataset with different edge density levels from the graph $\mathcal G(\mathcal W, \mathcal E)$. This simulation is run on $|\mathcal W|=14$ and $|\mathcal W|=28$ and the value of $|\mathcal S_p|$ is set to~$5$.

The result of this simulation for $|\mathcal W|=14$ shows that, for the leader-based approach, the performances for all metrics get higher with the increase of the edge density. For instance, the uncertainly level on the selected team skill decreases and tends to zero when the edge density reaches one. This can be explained by the fact that, by increasing the edge density of the graph, the number of hops between the leader and any worker in $\mathcal G(\mathcal W, \mathcal E)$  decreases. Moreover, the number of directly connected workers increases until the resultant graph is fully connected (i.e., everyone knows everyone). Also, the team skills level, budget allocation, and relationships rate increase when increasing the edge density because the team leader will have more workers connected to it and consequently more vast choices in its vicinity. When increasing the edge density, the relationships rate within the team increases because team members are more likely to have more connections. However, the performances of the platform-based approach, except the relationships rate, remains basically invariant while varying the edge density. We notice a growth in the relationships rate and a slight decrease in all the other metrics. This can be explained by the fact that the platform is basing its recruitment decision on the workers' history and any changes of the edges density in $\mathcal G(\mathcal W, \mathcal E)$ will only affect the relationships term in the objective function. Notice that, for the leader-based approach, the recruited team skills level exceeds the one of the platform-based approach when the SN network becomes nearly fully connected. These observation remains valid for $|\mathcal W|=28$. In fact, the effect of expanding the network while maintaining the number of required team members enhanced slightly the performance of both recruitment strategies by increasing the team skill level and decreasing the recruiter uncertainty. We notice that, for the leader-based approach, the team skill level curve has a higher slop and converge faster than when  $|\mathcal W|=14$.

In Fig.~\ref{2figures}, we present an example of the recruited teams using both recruitment strategies for $|\mathcal W|=14$, $|\mathcal S_p|=4$. The figure shows that the leader-based approach recruits a congregated team (workers are close to each other in the SN) while~the platform-based~approach~recruits~a~team~relatively scattered~but~with~higher~skills.

\begin{figure}[t!]
    \centering
    \vspace{0.25cm}
    \hspace{-0.35cm}
    \includegraphics[width=8.9cm]{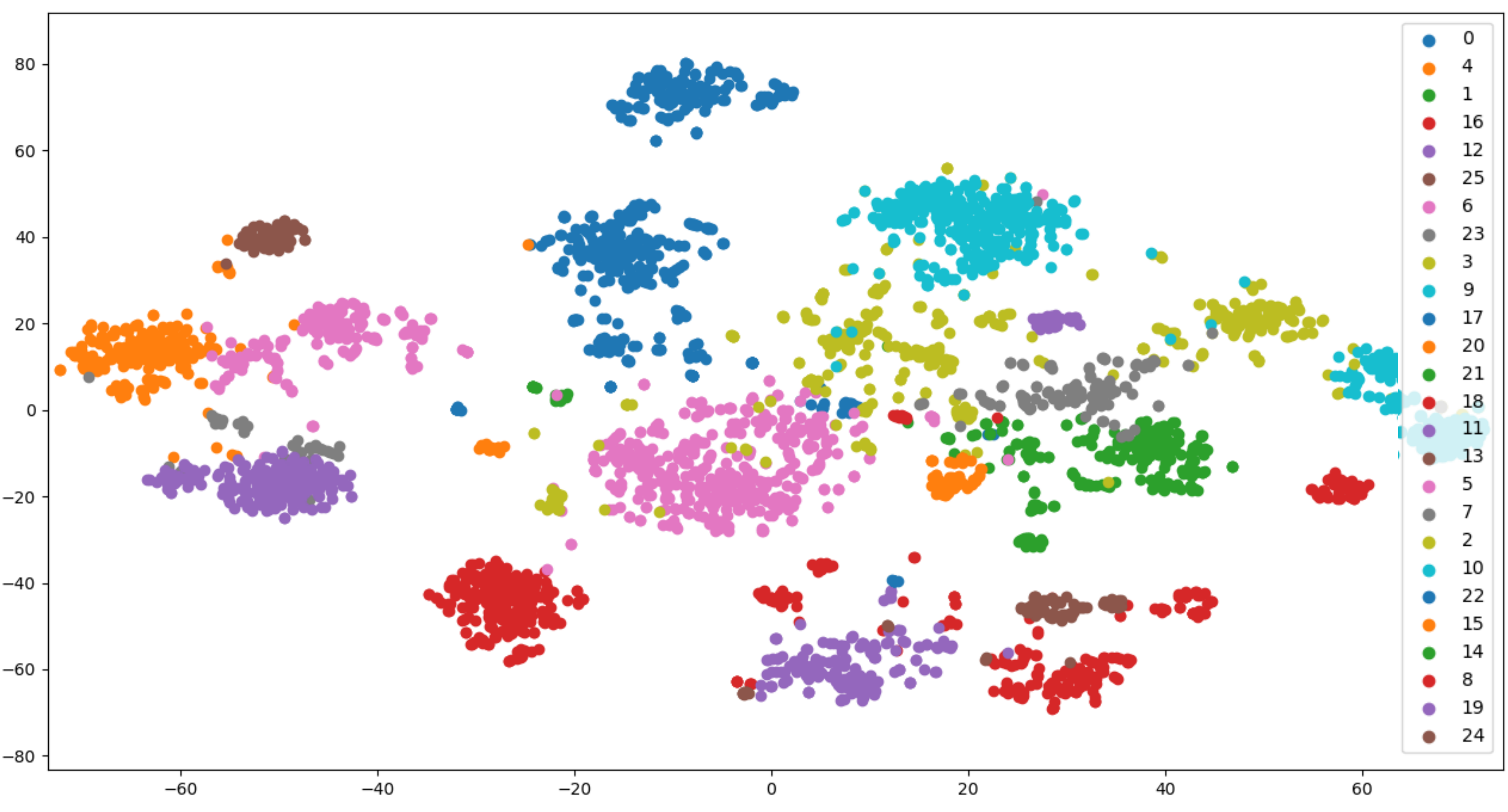}
\caption{2-D representation showing the result of the T-SNE  and GEMSEC for embedding and self-clustering where colors represent workers belonging to the same  cluster. The legend indicates the color and index of each cluster.}
\label{embed1}
\end{figure}

\begin{figure}[t!]
    \centering
    \vspace{0.25cm}
    \hspace{-0.35cm}
    \includegraphics[width=9cm]{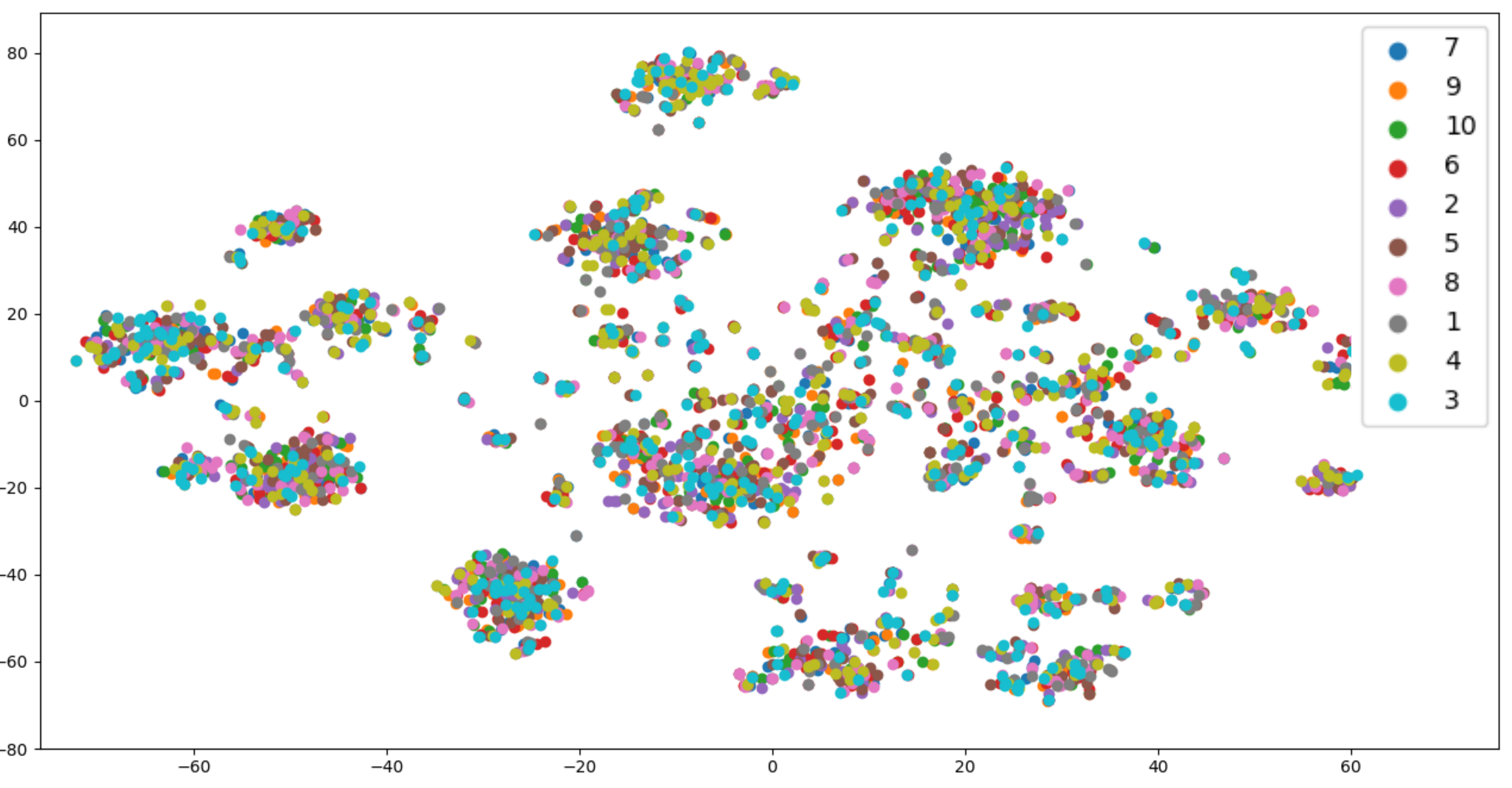}
\caption{2-D representation showing the result of the T-SNE using the GEMSEC for embedding and self-clustering where colors represent the dominant skills of each worker. The legend indicates the color and index of each possible skill.}
\label{embed2}
\end{figure}

\begin{figure}[t!]
    \centering
      \vspace{0.12cm}    \hspace{-0.35cm}
    \includegraphics[height=5cm,width=9cm]{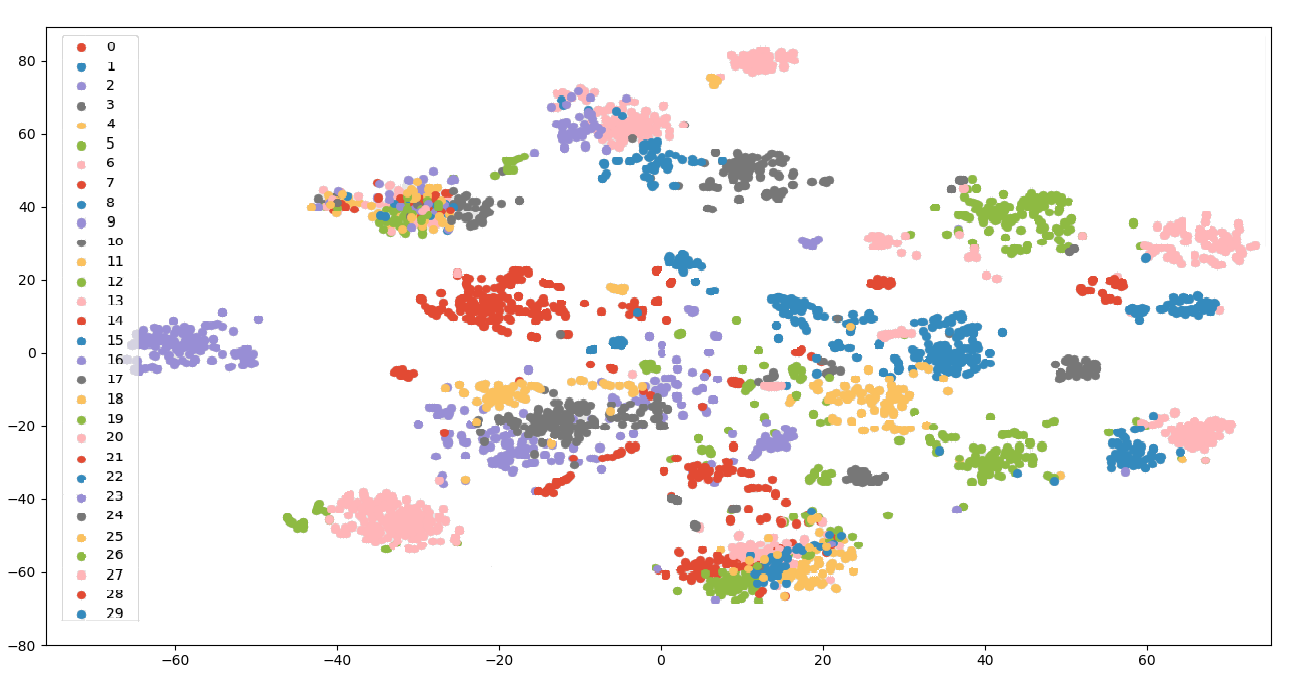}
\caption{Result of Edge-Attribute embedding using the T-SNE, ASNE for embedding, and GEMSEC for clustering where colors represent workers belonging to the same cluster. The legend indicates the color and index of each cluster.}
\label{embed3}
\end{figure}

\subsection{Graph Embedding and Clustering Performances}
Before comparing the results of the ILP and the proposed low-complexity algorithms, we propose to analyze the performance of the employed graph embedding techniques and clustering algorithms with the studied dataset (i.e., Phase I till Phase III of the proposed recruitment approach). After performing the node embedding on the SN graph using GEMSEC, we obtain a 23-D vector. We run T-SNE as the dimensionality reduction algorithm to speed up the computations. Next, we use the output of the T-SNE as input to clustering component in the GEMSEC. The final output of this process is illustrated in Fig.~\ref{embed1} and Fig.~\ref{embed2}. As we can notice, in Fig.~\ref{embed1}, we color the nodes with different colors where each color represents the cluster color. In Fig.~\ref{embed2}, we color the same nodes but this time, the color reflects the skill type of each worker/node. As it is showing, each cluster represents a community that have multiple workers with diversified skills. The clustering effect is illustrated using the SN graph as showing in Fig.~\ref{embed1} and Fig.~\ref{embed2}. Note that this approach serves also as a community detection technique.

\begin{table}[!]
\begin{center}
\caption{\label{table2} Statistics showing the clustering efficiency of the two embedding/clustering algorithms: Phase I to Phase III}
\addtolength{\tabcolsep}{-0pt} \scalebox{1}{
\begin{tabular}{c|c|c|}
\cline{2-3}
 & \begin{tabular}[c]{@{}c@{}}Edge-only \\ Embedding\end{tabular} & \begin{tabular}[c]{@{}c@{}}Edge-Attribute\\ Embedding\end{tabular} \\ \hline
\multicolumn{1}{|c|}{Modularity}                                                                 & 0.632 & 0.314 \\ \hline
\multicolumn{1}{|c|}{Number of clusters}                                                         & 25    & 36    \\ \hline
\multicolumn{1}{|c|}{\begin{tabular}[c]{@{}c@{}}Avg number of \\ users per cluster\end{tabular}} & 160   & 111   \\ \hline
\multicolumn{1}{|c|}{\begin{tabular}[c]{@{}c@{}}Max number of\\ users per cluster\end{tabular}}  & 421   & 197   \\ \hline
\multicolumn{1}{|c|}{\begin{tabular}[c]{@{}c@{}}Min number of\\ users per cluster\end{tabular}}  & 253   & 6     \\ \hline
\end{tabular}}
\end{center}
\end{table}

After performing the Edge-Attribute embedding on the SN graph and their attributes (namely skill level and cost), we obtain also a 48-D feature vector.  We also run the dimensionality reduction algorithm (T-SNE) then perform clustering using GEMSEC. The final output of this process is illustrated in Fig.~\ref{embed3}. As we can see, in Fig.~\ref{embed3}, we color the nodes with different colors where each color represents the cluster color. This type of embedding has made it possible to cluster nodes based on their attributes and SN degree. In each of these cluster, the nodes are practically similar in term of attributes, and they are close in the SN graph. Therefore, each cluster represents an expertise. Some metrics highlighting the difference between the two clustering approaches have been computed and included in Table~\ref{table2}. As it is shown, the modularity using the edge-only embedding is higher than the modularity of the edge-attribute embedding. This can be explained by the fact that the first approach uses the node's edges only to compute the clusters and therefore, the resultant clusters contain nodes that are more close in the SN.

Note that, for dimensionality reduction, we could have used the famous PCA. However, PCA is a linear algorithm and therefore it will not be able to interpret the complex polynomial relationship between features in the SN while t-SNE is made to capture exactly that. Also, PCA and t-SNE are not mutually exclusive. When dealing with highly dimensional data where t-SNE simply does not scale. We may use PCA first to reduce the dimensionality of the data and then, taking the top principal components, we apply t-SNE for visualization.

\begin{figure}[t]
\hspace{-0.0cm}
       \includegraphics[width=9cm]{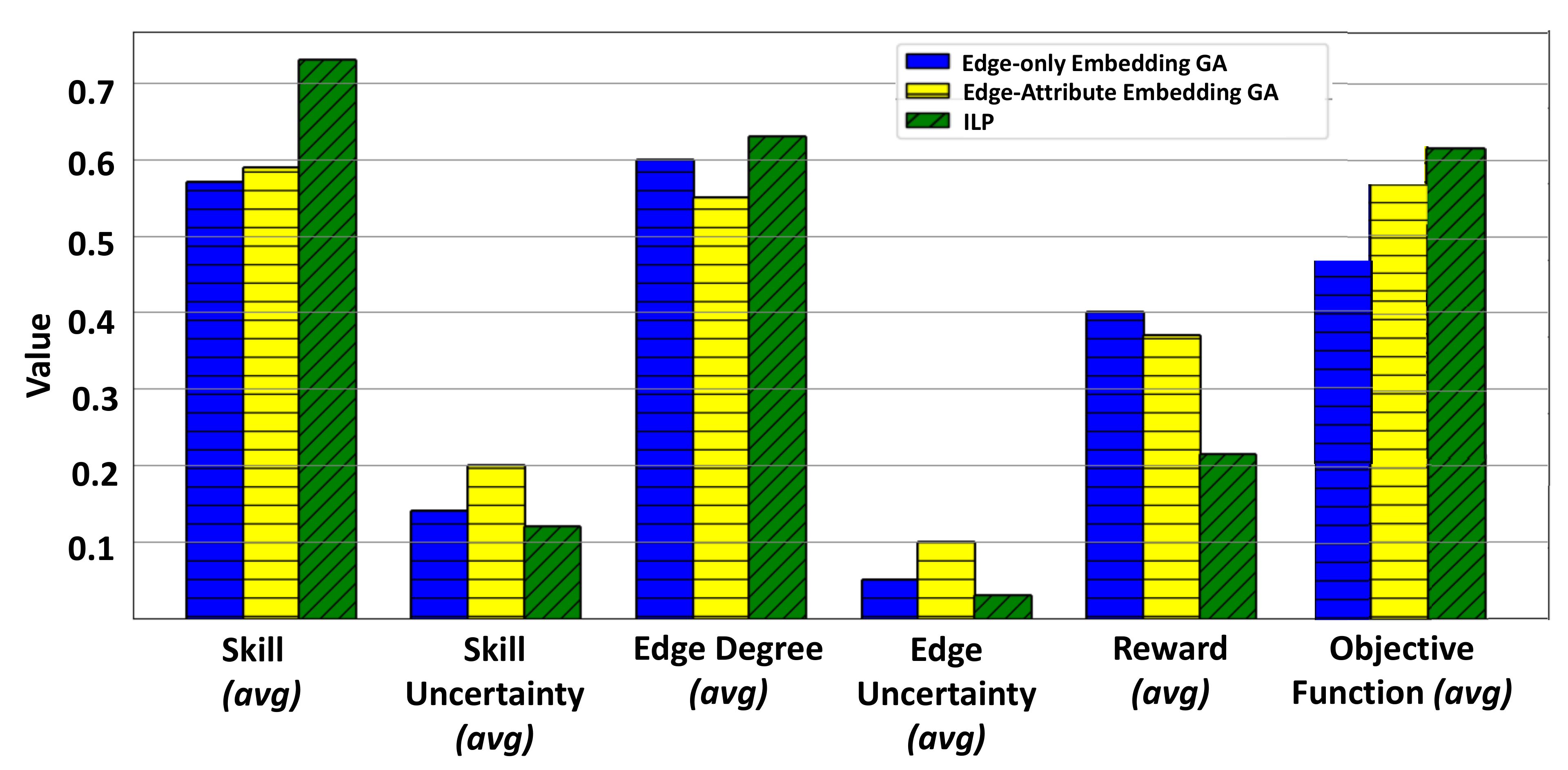}
    \caption{Performances of the proposed CMCS recruitment approaches using $|\mathcal W|=14$. The Edge-only embedding and Edge-Attribute embedding GA-based algorithms versus the optimal ILP.}    \label{allalgos}
    \vspace{-0.0cm}
\end{figure}

\subsection{Proposed Algorithm Performance Analysis}
In this subsection, we evaluate the performances of the proposed algorithm against the optimal ILP. For simplification purposes, and because extensive evaluations were conducted between the two recruitment strategies using the ILP, the proposed CMCS recruitment algorithm was only implemented for the platform-based scheme. However, theoretically, the results remain valid even for the leader-based scheme. For the performances analysis, we run a Monte-Carlo simulation where $1,000$ realizations of different project settings are generated for the Facebook dataset. For the GA, we define $N_{b}$ as the base population size, $N_{i}$ as the number of iterations, $R_{c}$ as the rate of crossover, and $R_{m}$ as the rate of mutation. we set $N_b=1000$, $N_i=500$, $N_c=0.4$ and $N_m=0.8$.
The choice of these values was carefully picked based on several conducted trials. \textcolor{black}{From the initial $4039$ pool of workers in the dataset, we pick at each realization only $|\mathcal W|=14$ workers. This is done to ensure fairness comparison between the performances of the ILP and the proposed approach as the ILP has an exponential running time and consequently, it fails to converge in a reasonable time for higher values of $|\mathcal W|$ . Furthermore, the ILP was run on $14$ pool of workers. However, for our proposed solution, and by the means of the means of the embedding and clustering techniques, this initial pool of workers is reduced to only the best potential candidates that are estimated to have the best performances and the GA is run on the resultant shrunk pool.} We evaluate the performances of the ILP, a GA algorithm based on Edge-only embedding, and a GA based on Edge-Attribute embedding using the recruitment key metrics: skills, skills uncertainty, edge degree, edge uncertainty, and the reward. For each of the $1,000$ simulation, we pick random $|\mathcal W|$ workers from the total available $4039$ that satisfies the following constraints: i) for the edge-only embedding, $|\mathcal W|$ must result in at least $|\mathcal S_p|$ clusters, ii) for the edge-attribute embedding, $|\mathcal W|$ must result at least in one cluster containing at least $|\mathcal S_p|$ nodes.  If the selected $W$ workers validate these conditions, we proceed and perform the worker selection phase. As Fig.~\ref{allalgos} illustrates, we notice that the performances of the proposed CMCS recruitment algorithm for the attributed embedding achieves better results than the Edge-only embedding with close performances to the optimal ILP.

\begin{figure}[t]
\hspace{-0.2cm}
        \includegraphics[width=9cm]{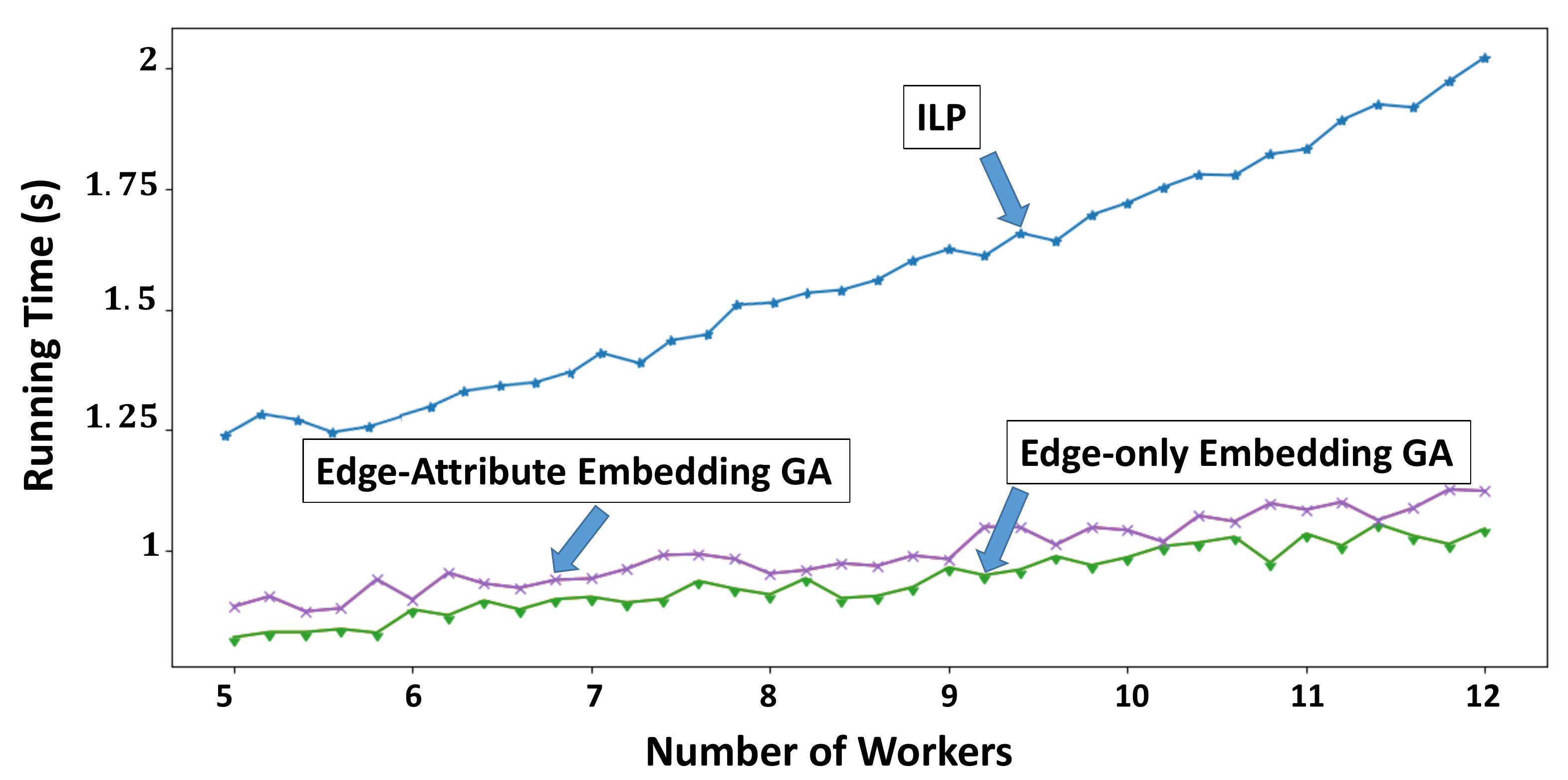}
    \caption{Running time in seconds for the Edge-only embedding and Edge-Attribute embedding GA-based algorithms and the optimal ILP.}
    \label{runningtime}
    \vspace{-0.0cm}
\end{figure}

\begin{figure}[t]
\centering
        \includegraphics[height=5cm,width=8.75cm]{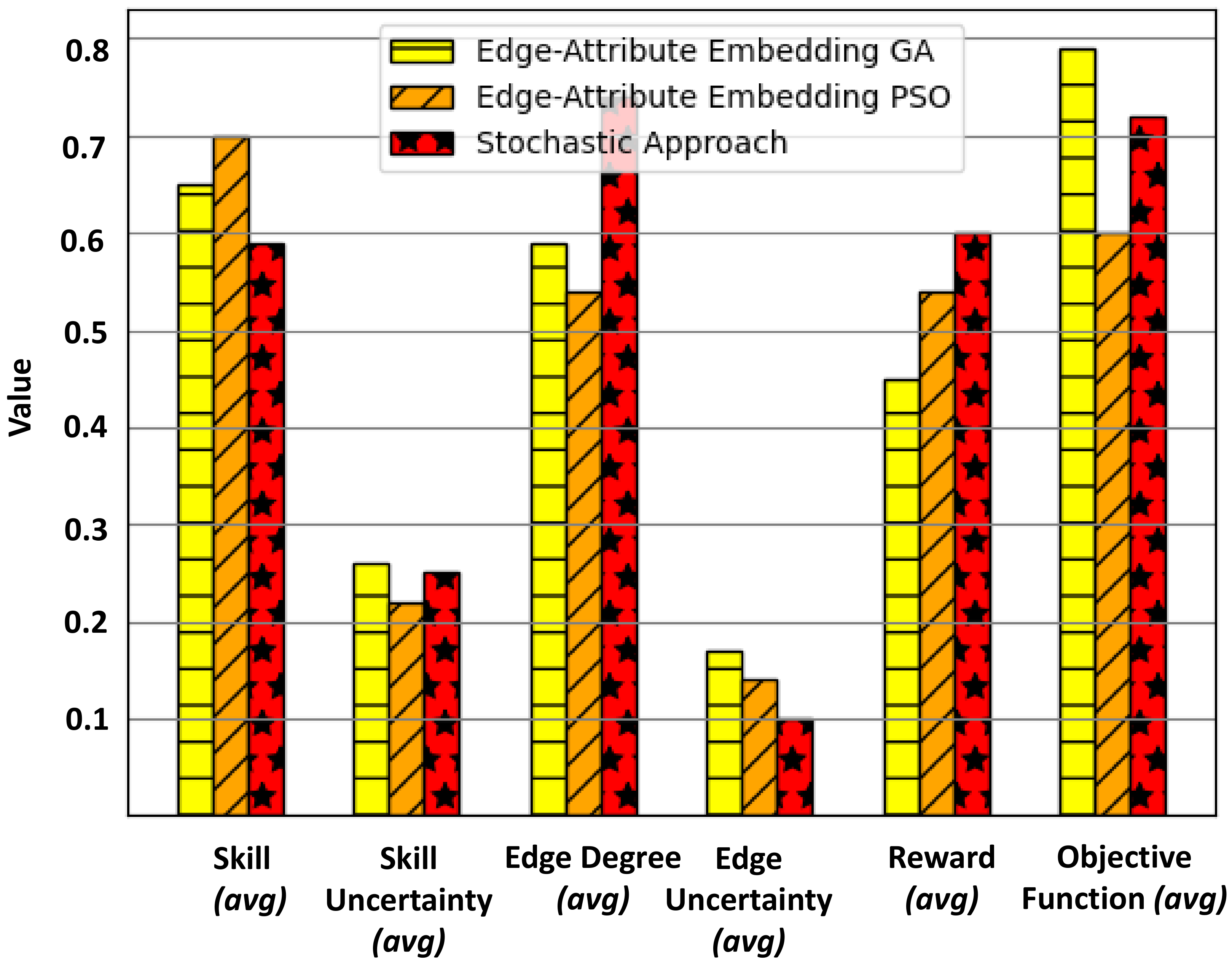}
        \caption{\textcolor{black}{Performances of the proposed CMCS recruitment approaches using $|\mathcal W|=1800$. The Edge-Attribute embedding GA-based algorithm versus the Edge-Attribute embedding PSO-based algorithm and CMCS Stochastic approach.}}
    \label{allalgos2}
    \vspace{-0.0cm}
\end{figure}

\begin{figure}[t]
\hspace{-0.2cm}
        \includegraphics[height=5cm,width=8.75cm]{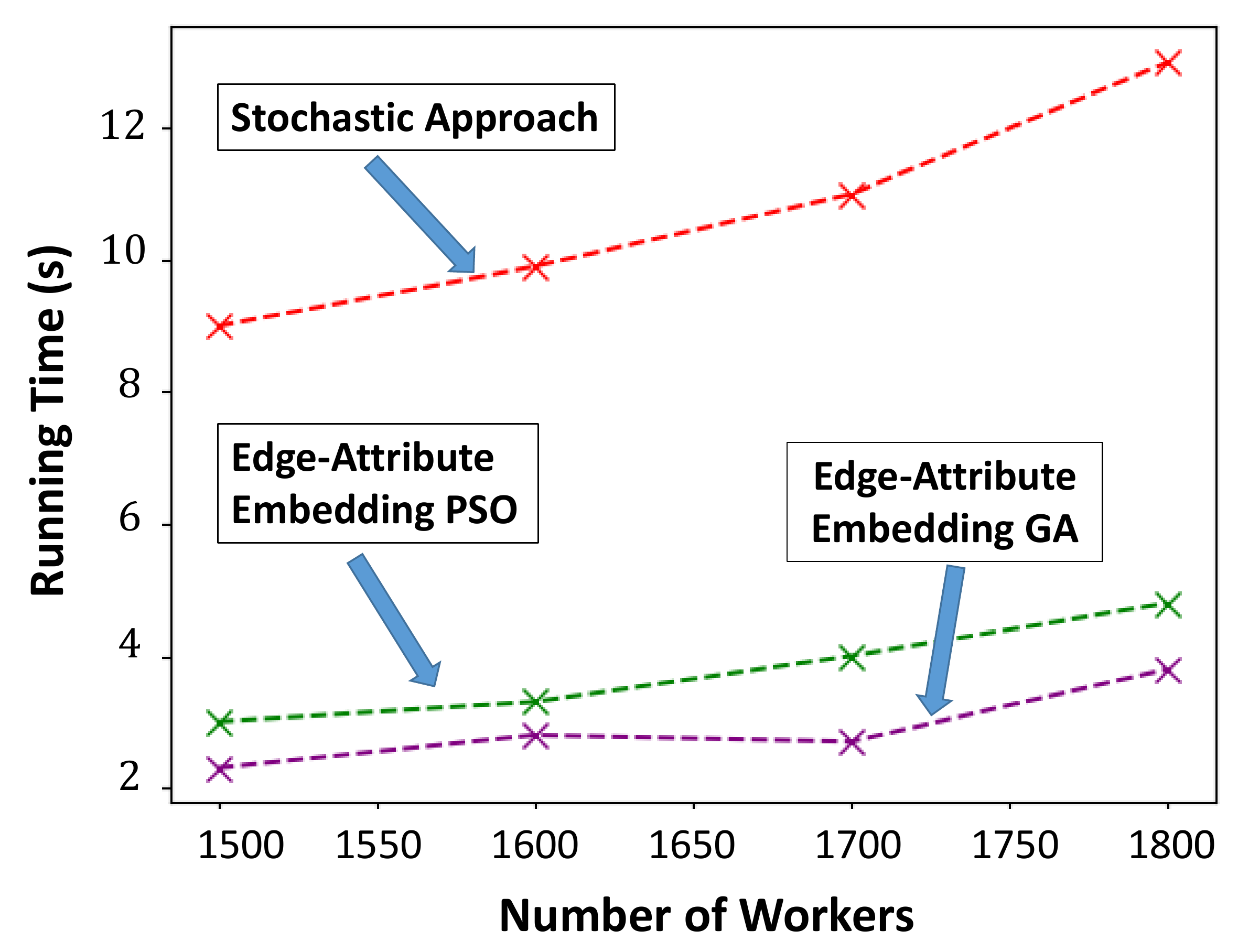}
    \caption{\textcolor{black}{Running time in seconds for the Edge-Attribute embedding GA-based algorithm, the Edge-Attribute embedding PSO-based algorithm, and CMCS Stochastic approach.}}
    \label{runningtime2}
    \vspace{-0.0cm}
\end{figure}

In order to evaluate the time complexity of the CMCS algorithms: the proposed low-complexity CMCS recruitment approach along with the optimal algorithm, we perform a running time simulation during which we vary the number of available workers in the platform and compute the running time needed for the three algorithms to converge. As Fig.~\ref{runningtime} shows, we notice that the optimal approach has the highest running time among the three algorithms and that the time complexity gap increases while increasing the number of available workers. Also, the optimal ILP has an exponential running time because the running time increases exponentially with the size of the input (i.e., number of workers). However, we note that the time complexity of the proposed CMCS recruitment approach with the two embedding variation is polynomial. The GA algorithm based on Edge-Attribute embedding has a slightly higher running time than the Edge-only Embedding GA.

\textcolor{black}{In another experiment, we increase the pool of initial workers to $|\mathcal W|=1800$ and evaluate the performances of the proposed Edge-Attribute Embedding GA with two other CMCS recruitment algorithms. The first is similar to the proposed approach but instead of using GA, we employ the PSO algorithm. We refer to this algorithm as "Edge-Attribute Embedding PSO". Hence, the workers of the network undergo an attribute embedding and clustering to shrink the pool of workers before running the PSO algorithm. The second approach, identified as the "Stochastic approach", is a probabilistic technique that uses the optimal stopping strategies. It is based on the odds algorithm~\cite{8752756,9021910}. The purpose of this Monte-Carlo simulation is to evaluate the recruitment key metrics: skills, skills uncertainty, edge degree, edge uncertainty, and the reward for each of these three algorithms for a large-scale network. As performed previously in Fig.~\ref{allalgos}, for each of the $1,000$ simulations, we pick random $|\mathcal W|=1800$ workers from the total available $4039$.  We notice that the Edge-Attribute Embedding PSO achieves, in certain key metrics, higher performances than the Edge-Attribute Embedding GA. However, the highest overall objective function, describing the efficiency of the recruited team was realized by the Edge-Attribute Embedding GA. The Stochastic approach performs poorly compared to the other two algorithms. This might be due to the fact that it operates on the entire $|\mathcal W|=1800$ workers and does not undergo search space reduction. We also perform time complexity analysis for these three algorithms while varying the number of workers between $1500$ and $1800$. As shown in Fig.~\ref{runningtime2}, the Edge-Attribute embedding GA algorithm has a slight lower running time than the Edge-Attribute embedding PSO. The stochastic approach yields an elevated running time surpassing $10s$ per simulation.}

\section{Conclusion and Future Work}\label{sec6}

In this paper, we proposed a Collaborative Mobile Crowdsourcing (CMCS) recruitment approach in large-scale IoT networks. The objective is to form a skilled and socially connected team that matches the project requirements submitted to the platform by a task requester. We first formulated the CMCS team formation problem as an Integer Linear Program (ILP) that optimally recruit workers and form teams that satisfy the required project skills while being socially connected. We also proposed two recruitment strategies. The first proposed strategy is a platform-based approach which exploits the platform knowledge to form the team. The second one is a leader-based approach that designates one of the workers as a leader of the team and based on its knowledge about their SN neighbors the team is recruited.

Due to the high computational complexity of the ILP, we proposed a low-complexity heuristic recruitment approach that uses the network embedding/clustering techniques and rely on the genetic algorithm as a solver. This model proves to have lower overhead with close performances to the optimal ILP. Results of the effectiveness and efficiency of the proposed heuristic CMCS recruitment algorithm have been verified via experiments on the Facebook dataset. It is shown that the proposed graph neural network based approach is able to shrink the search space of the CMCS dataset into small clusters of workers sharing common skills and/or social relations, which allow its easy application on large-scale CMCS platforms. 

Based on the findings of this paper, our future work will treat the privacy concerns that workers could encounter when revealing their social interactions. Also, we will consider an adaptive feedback recruitment mechanism assuring submission quality control where the quality of the responses and collaborative behaviors among teams are assessed. \textcolor{black}{Furthermore, we will enhance the learning component in our proposed approach, in terms of convergence speed and accuracy, by incorporating Bayesian learning in the embedding optimization and harnessing the power of inductive learning to avoid repeating the mapping process when new nodes are added to the social IoT network. }
\bibliographystyle{IEEEtran}
\bibliography{references}

\begin{IEEEbiography}
[{\vspace{0.1cm}\includegraphics[width=1.1in,height=1.34in]{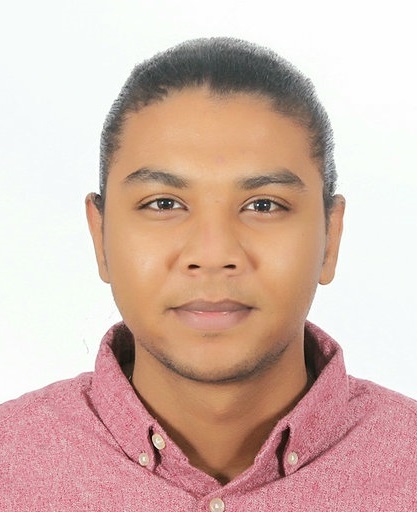}}]
{AYMEN HAMROUNI}
(Student Member, IEEE) received the Diplome d’Ingenieur degree (\textit{with summa cum laude}) in Telecommunication engineering from the Ecole Superieure des Communications de Tunis (SUP’COM), Tunis, Tunisia, in 2019. In 2019, he was a Research Assistant with the Stevens Institute of Technology, Hoboken, NJ, USA. He is currently a Research Scholar with the School of Systems and Enterprises, Stevens Institute of Technology. His research interests include the intersection of mobile crowdsourcing, applied machine learning, optimization, social network analysis, mathematical modeling, graph theory, and the Internet-of-Things.
\end{IEEEbiography}
\vspace{-0.7cm}
\begin{IEEEbiography}
[{\vspace{0.3cm}\includegraphics[width=1.1in]{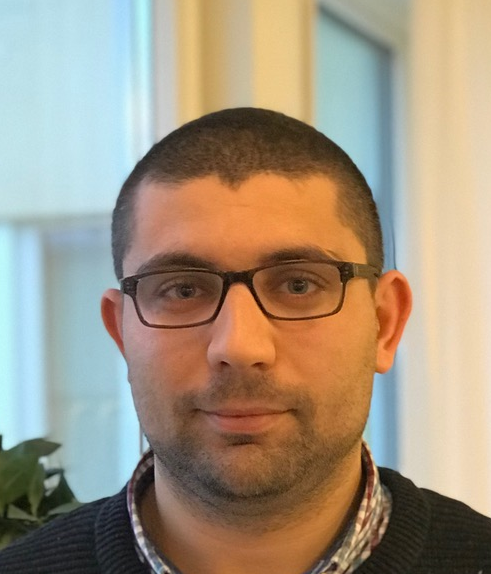}}]%
{HAKIM GHAZZAI} (Senior Member, IEEE) received the Ph.D. degree in electrical engineering from KAUST, Saudi Arabia, in 2015, and the Diplome d'Ingenieur (Hons.) in telecommunication engineering and the master's degree in high-rate transmission systems from the Ecole Superieure des Communications de Tunis (SUP'COM), Tunis, Tunisia, in 2010 and 2011, respectively. He was a Visiting Researcher with Karlstad University, Sweden, and a Research Scientist with the Qatar Mobility Innovations Center (QMIC), Doha, Qatar, from 2015 to 2018. He is on the Editorial Board of the IEEE Communications Letters, the IEEE Open Journal of the Communications Society, and Frontiers in Communications and Networks. He has authored over 130 articles in peer-reviewed journals and conferences. He is currently a Research Scientist with the Stevens Institute of Technology, Hoboken, NJ, USA.  His general research interests are on the areas of wireless networks, UAVs, the Internet of Things, intelligent transportation systems, and optimization.
\end{IEEEbiography}
\vspace{-0.7cm}

\begin{IEEEbiography}
[{\vspace{0.1cm}\includegraphics[width=1.1in,height=1.34in]{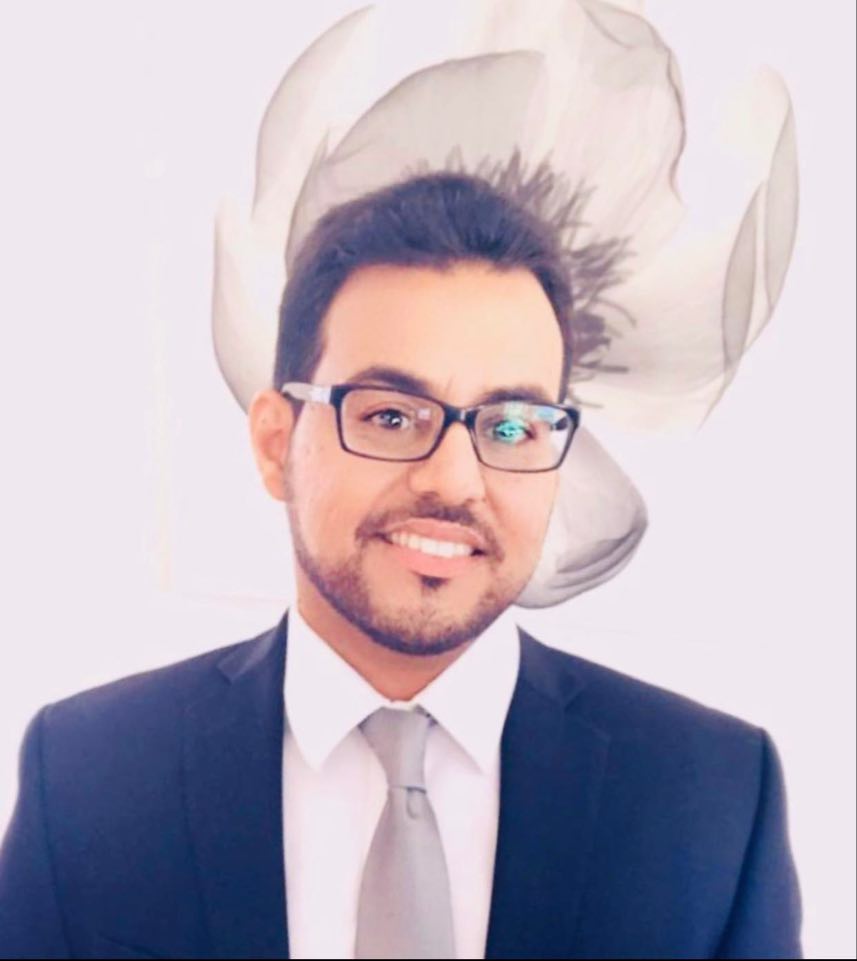}}]%
{TURKI ALELYANI} (Member, IEEE) received his Ph.D. in Software Engineering from Stevens Institute of Technology, New Jersey, United States. Prior to that, he received his master degree in Computer Science from Stevens as well. Dr. Alelyani's research studies Socio-Technical Systems Design in order to overcome some of the challenges in motivation, engagement, coordination and collaboration. His research is applied into different domains including engineering design, Software Engineering, and social computing. He approaches this research by conducting empirical studies which involve using statistical analysis, machine learning, and experimental design.

\end{IEEEbiography}
\vspace{-0.7cm}

\begin{IEEEbiography}
[{\vspace{0.3cm}\includegraphics[width=1.1in]{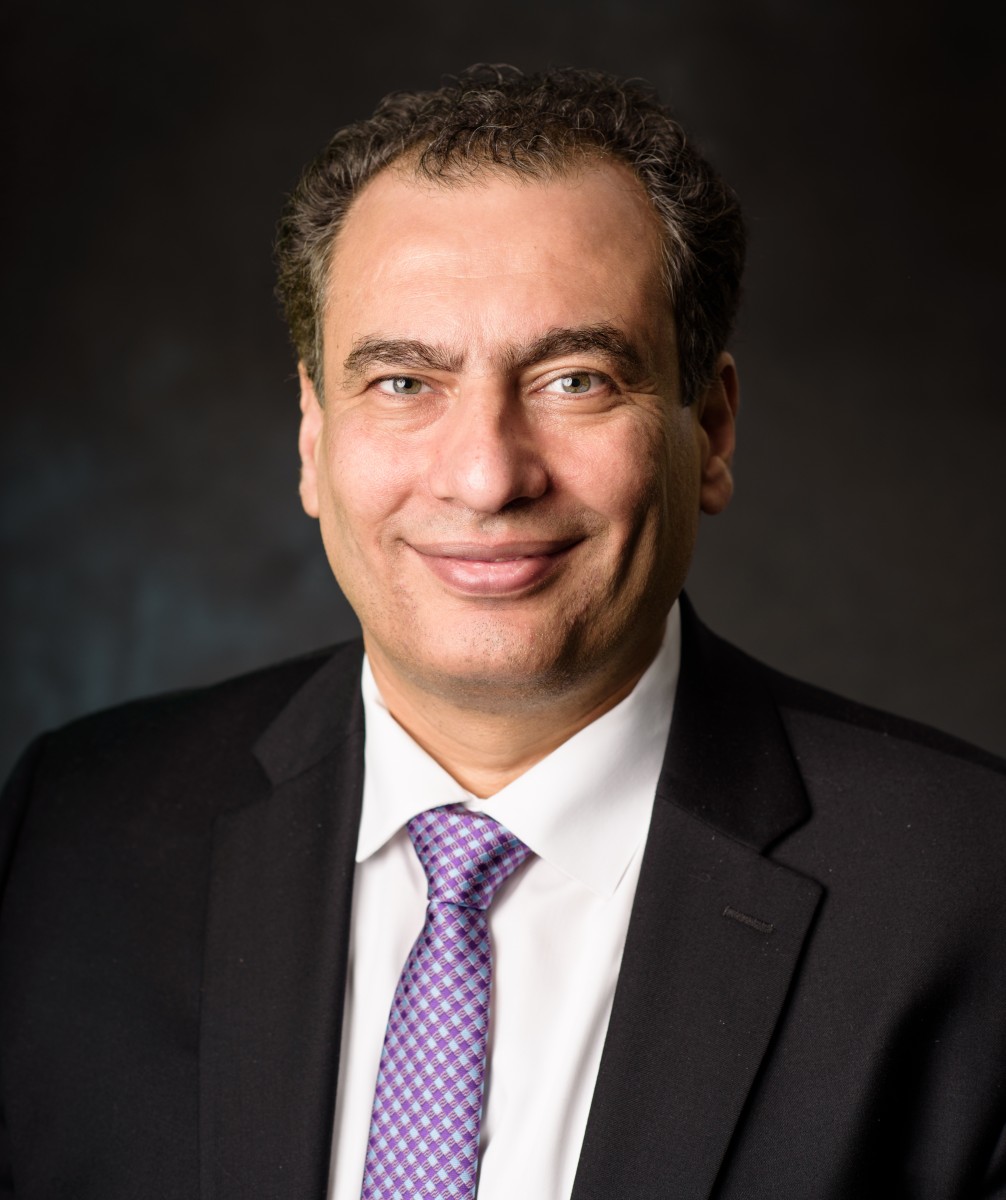}}]%
{YEHIA MASSOUD} (Fellow Member, IEEE) received the Ph.D. degree in electrical engineering and computer science from the Massachusetts Institute of Technology, Cambridge, MA, USA. He is currently the Dean of the School of Systems and Enterprises, Stevens University of Science and Technology, Hoboken, NJ, USA. He has authored over 325 articles in peer-reviewed journals and conferences. He is a Fellow of the IEEE and has served as a Distinguished Lecturer by the IEEE Circuits and Systems Society and as an elected member of the IEEE Nanotechnology Council. He was selected as one of ten MIT Alumni Featured by MIT's Electrical Engineering and Computer Science department in 2012. He was a recipient of the Rising Star of Texas Medal, in 2007, the National Science Foundation CAREER Award, in 2005, the DAC Fellowship, in 2005, the Synopsys Special Recognition Engineering Award, in 2000, several best paper award nominations. Dr. Massoud has held several academic and industrial positions, including a member of the technical staff with the Advanced Technology Group, Synopsys, Inc., CA, USA, a tenured faculty with the Departments of Electrical and Computer Engineering and Computer Science, Rice University, Houston, USA, the W. R. Bunn Head of the Department of Electrical and Computer Engineering at UAB, Birmingham, USA, and the Head of the Department of Electrical and Computer Engineering, Worcester Polytechnic Institute, USA. Massoud has served as the editor of Mixed-Signal Letters—The Americas and also as an associate editor of IEEE Transactions on Very Large Scale Integration Systems and IEEE Transactions on Circuits and Systems I. 
\end{IEEEbiography}
\end{document}